\documentclass[conference]{IEEEtran}

\usepackage{amsmath,amssymb}
\usepackage{graphicx}
\usepackage{booktabs}
\usepackage{xcolor}
\usepackage{tikz}
\usetikzlibrary{arrows.meta,shapes,positioning,fit,backgrounds}
\usepackage{hyperref}
\usepackage{url}
\usepackage{cite}
\usepackage{array}
\usepackage{tabularx}
\usepackage{multirow}
\usepackage{enumerate}

\hypersetup{colorlinks=true, linkcolor=blue!50!black, citecolor=blue!50!black,
            urlcolor=blue!50!black}

\newcommand{\eod}{\text{EOD}}
\newcommand{\ece}{\text{ECE}}
\newcommand{\mce}{\text{MCE}}

\begin{document}

% ============================================================
%  TITLE
% ============================================================
\title{KAISEN: Reproducible Subgroup Fairness Auditing\\for Clinical Risk Models}

\author{
\textbf{Sparsh Roy}$^{1,2}$,
\textbf{Samuel Girmachew}$^{2}$,
\textbf{Nishita Chavan}$^{3}$\\[6pt]
$^{1}$Massachusetts Institute of Technology\\
$^{2}$Hopewell Valley Central High School\\
$^{3}$East Brunswick High School\\[6pt]
\texttt{sparshr@mit.edu, samigirmachew@gmail.com, nishitac0408@gmail.com}
}
\maketitle

% ============================================================
%  ABSTRACT
% ============================================================
\begin{abstract}
Clinical risk models routinely achieve strong aggregate performance while
producing materially different error rates across patient subgroups.  Audit
pipelines are proposed to catch this, but their components are rarely
stress-tested, so it is unclear which parts of an audit can be trusted and
under what conditions.  We present KAISEN, a five-phase audit pipeline
covering subgroup stratification, disparity measurement, mechanism
diagnostics, post-hoc mitigation, and longitudinal drift monitoring, and we
evaluate each phase to the point of failure on a synthetic benchmark of 16
disease tasks, 15 social-determinant axes drawn from Healthy People
2030~\cite{hp2030}, and three prespecified intersections.  Four findings
follow.  (i)~Significance tracks each axis's gap measured against its own
minimum detectable effect.  Across the 15 single axes, rank correlation
between significance count and raw equalized-odds difference (EOD) is
$\rho=0.56$; standardizing EOD by that detectable floor raises it to
$\rho=0.78$.  (ii)~Per-group
threshold optimization reduces EOD in 48 of 48 held-out runs (paired
$\Delta=-0.285$, 95\%~CI $[-0.313,-0.252]$), whereas group-wise Platt scaling
is the better calibrator yet behaves as a coin flip on EOD (19 of 48 runs
improved, 95\%~CI $[0.26,0.55]$).  Its mean EOD effect is indistinguishable
from zero, so what an audit should report about it is the variance.
(iii)~The mechanism diagnostic classifies 144 of 144 controlled cases
correctly but recovers none of 48 model-driven cases when its proxy columns
are misspecified, and emits no signal that it has failed.  (iv)~CUSUM
detection failures and false alarms track the cohort realization far more
closely than the disease: at the reference threshold, all 27 false alarms and
7 of 8 missed shifts come from different seeds ($\chi^2$ $p=0.002$).  An alarm
threshold tuned on one cohort's in-control stream therefore fails to
transfer.  All results are on synthetic data with known ground truth and
do not establish clinical validity.  Code, artifacts, and the scripts that
emit every number in this paper are released.
\end{abstract}


% ============================================================
%  I. INTRODUCTION
% ============================================================
\section{Introduction}

Aggregate performance can hide clinically relevant failures.  A risk model may
retain strong overall AUROC while its false-negative rate, calibration, or
ranking performance differs sharply across patient groups.  These gaps can
arise without using a protected attribute directly, because electronic health
records encode proxies for access, utilization, neighborhood conditions, and
prior care.  They can also widen after deployment as patient populations and
coding practices shift.

Most evaluation pipelines are not designed to surface this.  They report
aggregate discrimination and calibration, add subgroup analyses
inconsistently, and rarely track fairness metrics over time.  A model can pass
a conventional validation review while producing different errors for patients
who differ by insurance status, income, or neighborhood deprivation.

A growing number of audit frameworks address this gap.  We take up a prior
question about those frameworks.  An audit is a measurement instrument, and an
uncharacterized instrument fails quietly.  A clean audit result admits at
least three readings: the disparity really is absent, the diagnostic was
aimed at the wrong features, or the subgroup was too small for the test to
resolve anything; the printed output looks identical in every case.  This
paper asks \emph{under what conditions each component of an audit stops
working}, and whether the failure shows up in what the audit reports.

Answering that requires ground truth, so we build a synthetic benchmark in
which the disparity mechanism is known by construction and then break each of
the audit's assumptions in turn.  Interpretability is bought here at the cost
of external validity.  Every result below describes the audit procedure under
controlled conditions and says nothing about disparity in any clinical
population.

\subsection{Scope}
\label{sec:scope}

We evaluate binary risk prediction on 16 synthetic disease tasks: diabetes,
heart failure, sepsis, chronic kidney disease, stroke, 30-day readmission,
COPD, asthma, hypertension, depression, obesity, myocardial infarction,
pneumonia, acute kidney injury, atrial fibrillation, and in-hospital
mortality.  The audit covers race/ethnicity, sex, and age alongside 12
additional social-determinant (SDOH) axes, plus three prespecified
intersections.  Race and sex are included, but restricting an audit to those
fields misses disparities associated with insurance, income, education,
neighborhood deprivation, housing, and access~\cite{obermeyer2019}.

We make no claim to clinical validity.  Causal effects are outside what this
design can identify, which is why Phase~III is called a mechanism
\emph{diagnostic} throughout.  Nor do we recommend a deployment configuration
for any component: the operating points that work on this generator are
properties of the generator.

\subsection{Contributions}

\begin{enumerate}[(i)]
\item A five-phase audit pipeline with conditional permutation inference for
      equalized-odds disparities, specified so that each phase's assumptions
      are separately testable
      (\S\ref{sec:formulation}--\S\ref{sec:framework}).
\item An account of \emph{what subgroup significance actually tracks}.  Effect
      size and evidence rank axes differently; standardizing each axis's EOD
      by its own minimum detectable effect largely reconciles them, and the
      two axes that reach significance in no disease are the two with the
      smallest standardized effect
      (\S\ref{sec:support}, \S\ref{sec:evidence}).
\item A run-level comparison of two post-hoc mitigations, with paired
      uncertainty on the deltas, in place of the usual comparison of
      configuration means.  The better calibrator turns out to be the
      unreliable fairness intervention, and because its mean EOD effect is
      inseparable from zero, what we report is its variance
      (\S\ref{sec:mitigation}).
\item Stress tests that break the mechanism diagnostic's two assumptions and
      show that the more consequential failure is silent
      (\S\ref{sec:attribution}).
\item A characterization of a CUSUM fairness monitor across an alarm-threshold
      sweep, reported with in-control average run length, in which detection
      failures and false alarms track the cohort realization far more closely
      than the disease (\S\ref{sec:monitoring}).
\item An analysis of three deployed clinical decision contexts that derives
      which subgroup metric each one makes consequential, so the audit reports
      the metric the decision depends on
      (\S\ref{sec:clinical_context}, \S\ref{sec:usecases_results}).
\item A released synthetic SDOH benchmark generator spanning 16 disease
      parameterizations $\times$ 15 sensitive axes with correlated latent
      socioeconomic structure, the full audit codebase, and the scripts that
      regenerate every number in this manuscript from committed
      artifacts~\cite{kaiser_repo}.
\end{enumerate}

% ============================================================
%  II. RELATED WORK
% ============================================================
\section{Related Work}

\subsection{Fairness criteria in machine learning}

Formal fairness criteria for binary classifiers are
well-established~\cite{dwork2012,hardt2016,chouldechova2017,barocas2019}.
Demographic parity, equalized odds, and within-group calibration are the most
studied, and a central result is that they are mutually incompatible under
realistic class imbalance~\cite{chouldechova2017,kleinberg2016}.  KAISEN does
not try to satisfy them simultaneously.  It measures all of them, reports the
trade-off surface, and leaves the choice of which violations are clinically
unacceptable to the deploying institution.  Our Phase~IV results give a run-level empirical instance of that
incompatibility; we propose no resolution to it.

\subsection{Disparities in clinical risk prediction}

Subgroup failures are documented in dermatology AI~\cite{oakden2020}, pulse
oximetry~\cite{sjoding2020}, EHR-based readmission
prediction~\cite{obermeyer2019}, and chest radiograph classifiers across 14
pathology categories~\cite{seyyed_kalantari2021}.  Obermeyer et
al.~\cite{obermeyer2019} found that a widely deployed risk-stratification
algorithm assigned lower scores to Black patients than to comparably ill White
patients because it used health cost as a proxy for health need.  The
algorithm's aggregate metrics gave no sign of it.  Seyyed-Kalantari et
al.~\cite{seyyed_kalantari2021} found that underdiagnosis rates are highest for
intersectional minority subgroups even when overall AUC exceeds 0.90, which
motivates per-group AUC reporting alongside population-level metrics.  KAISEN
extends this audit approach across the full SDOH axis space and across 16
disease tasks simultaneously.
Table~\ref{tab:disparity_examples} summarizes what each study measured and
what it implies for the audit design.

\subsection{Social determinants of health in clinical AI}

The Healthy People 2030 framework~\cite{hp2030} organizes SDOH into five
domains: economic stability, education, health-care access, neighborhood and
built environment, and social and community context.  Insurance
status~\cite{obermeyer2019}, area deprivation~\cite{kind2014}, and housing
instability~\cite{decker2011} have each been linked to differential clinical
outcomes, but fairness audits rarely measure all domains in the same study.
We treat the 15 axes in our benchmark as a practical approximation of this
space.

\subsection{Calibration and statistical testing for subgroups}

Calibration and equalized odds answer different questions.  Reliability
diagrams and expected calibration error summarize probability accuracy,
extending the binned goodness-of-fit approach of the Hosmer--Lemeshow
test~\cite{hosmer1980} to modern binary classifiers, while AUC measures
ranking performance.  Equal-width binning is a known source of bias in ECE
estimates, and equal-mass and debiased variants have been
proposed~\cite{nixon2019}; we report both schemes and quantify their
disagreement in \S\ref{sec:calibration}.  EOD is defined within outcome
strata, so we use conditional randomization for its permutation test, and
Benjamini-Hochberg correction controls the false discovery rate across the
prespecified disease-axis hypotheses~\cite{bh1995}.

\subsection{Counterfactual and causal approaches to fairness}

Counterfactual fairness~\cite{kusner2017} requires that a prediction not
change when the sensitive attribute is intervened on under the true causal
model.  This is hard to enforce in EHR data because the causal graph is rarely
known.  We use a weaker model-internal diagnostic that shifts known synthetic
proxy columns toward a reference-group mean.  The generator exposes the
injected proxy mechanism, which makes this tractable; it should not be read as
a causal intervention on real patient data, and \S\ref{sec:attribution} shows
what happens when the proxy set is wrong.

\subsection{Debiasing methods}

Debiasing methods split into three families.  Pre-processing modifies training
data via class reweighting~\cite{kamiran2012} or resampling to equalize group
representation.  In-processing constrains training directly; adversarial
debiasing~\cite{zhang2018} adds a discriminator that penalizes
group-predictive features.  Post-processing adjusts predictions after
training, with group-wise Platt scaling~\cite{guo2017} and per-group threshold
optimization~\cite{hardt2016} the standard examples.  Our experiments use
inverse-propensity weights only when fitting the group-wise calibration
models; the base predictor is not retrained.  This is post-hoc weighted
calibration, not a pre-processing intervention on the training set.

\subsection{Longitudinal fairness monitoring}

Most audit literature evaluates a model at a fixed point in time.
Post-deployment distribution shift can degrade fairness properties even when
aggregate performance appears stable~\cite{giobergia2025}.  Sequential
change-point methods such as CUSUM~\cite{page1954} detect when a tracked
statistic has shifted beyond a controlled false-alarm rate, and are
conventionally characterized by in-control average run length (ARL$_0$)
alongside detection delay~\cite{montgomery2009}.  We apply CUSUM to
subgroup-stratified EOD streams and report both quantities.  Little work
applies these methods to clinical AI fairness, and we treat ours as an initial
probe, well short of a validated clinical monitoring tool.

\begin{table*}[t]
\caption{Published examples of subgroup performance gaps in deployed clinical
  AI.  The \emph{disparity found} column reports what each study measured; the
  \emph{proposed mechanism} and \emph{implication} columns are our
  interpretation and are not claims established by the cited work.}
\label{tab:disparity_examples}
\centering
\begin{tabular}{p{2.6cm}p{1.9cm}p{2.3cm}p{3.6cm}p{2.4cm}p{1.9cm}}
\toprule
\textbf{Study} & \textbf{Clinical task} & \textbf{Groups compared} &
\textbf{Disparity found (published)} & \textbf{Proposed mechanism
(interpretation)} & \textbf{Implication for KAISEN} \\
\midrule
Obermeyer et al., \textit{Science} 2019~\cite{obermeyer2019} &
  Health-risk stratification (readmission / utilization) &
  Black vs.\ White (commercial insurance) &
  Equal-acuity Black patients scored $\approx$45\% lower; at equal risk score,
  Black patients were materially sicker &
  Cost-based proxy feedback loop: historical under-utilization by Black
  patients depresses spending labels used as ground truth &
  Ground-truth proxy choice encodes access inequity; KAISEN flags cost-proxy
  surrogates via audit of positive-rate disparity \\[4pt]
Sjoding et al., \textit{NEJM} 2020~\cite{sjoding2020} &
  Arterial oxygen saturation monitoring (ICU) &
  Black vs.\ White (two large academic hospitals) &
  Occult hypoxemia (SpO$_2{\ge}92\%$ despite SaO$_2{<}88\%$) was
  three times more frequent in Black patients &
  Melanin absorption artifact systematically inflates SpO$_2$ readings in
  patients with darker skin pigmentation &
  Input-level sensor bias can masquerade as model calibration error;
  KAISEN's MCE metric isolates worst-bin miscalibration where such
  artifacts concentrate \\[4pt]
Seyyed-Kalantari et al., \textit{Nat.\ Med.} 2021~\cite{seyyed_kalantari2021} &
  Chest X-ray pathology detection (14-class CheXpert / MIMIC-CXR) &
  Sex, insurance type, race, age $\ge$80 &
  Underdiagnosis rates consistently highest for intersectional minority
  subgroups across all 14 pathologies even when overall AUC exceeded 0.90 &
  Model trained to maximize population-level AUC; minority subgroups too
  small to influence loss gradient &
  High aggregate AUC masks subgroup underdiagnosis; KAISEN's per-group
  AUC and EOD surfaces this by design \\
\bottomrule
\end{tabular}
\end{table*}

% ============================================================
%  III. PROBLEM FORMULATION
% ============================================================
\section{Problem Formulation}
\label{sec:formulation}

\subsection{Data and notation}

Let $\mathcal{D}=\{(\mathbf{x}_i,a_i,y_i)\}_{i=1}^N$ be a labeled dataset
where $\mathbf{x}_i\in\mathcal{X}\subseteq\mathbb{R}^d$ is the feature vector
(lab values, vitals, coded history), $a_i\in\mathcal{A}=\{1,\ldots,K\}$ is a
sensitive attribute, and $y_i\in\{0,1\}$ is the binary clinical outcome.  A
trained model $f_\theta:\mathcal{X}\to[0,1]$ produces a risk score
$\hat{p}_i=f_\theta(\mathbf{x}_i)$, and a hard-threshold classifier at cutoff
$\tau$ produces $\hat{Y}_i=\mathbf{1}[\hat{p}_i\ge\tau]$.  We write
$\mathcal{D}_a=\{i:a_i=a\}$ for the index set of group $a$.

\subsection{Fairness criteria}

We evaluate four quantities simultaneously and do not claim to satisfy all of
them.  Each is defined for a reference pair $a,b\in\mathcal{A}$.

\textit{1) Equalized odds difference (EOD).}  Equalized odds requires equal
true positive rates (TPR) and equal false positive rates (FPR) across
groups~\cite{hardt2016}:
\begin{equation}
  \eod(a,b)=\max\!\bigl(|\text{TPR}_a-\text{TPR}_b|,\,
                        |\text{FPR}_a-\text{FPR}_b|\bigr).
  \label{eq:eod}
\end{equation}
Because the TPR term is estimated on positives only, the effective sample size
for EOD is the number of positive cases in the smaller cell, not its total
size.  \S\ref{sec:support} builds on this.

\textit{2) AUC disparity.}
\begin{equation}
  \Delta\text{AUC}(a,b)=|\text{AUC}_a-\text{AUC}_b|,
  \label{eq:auc_gap}
\end{equation}
where $\text{AUC}_a$ is the Wilcoxon--Mann--Whitney
estimator~\cite{delong1988} computed within subgroup $a$ only.

\textit{3) Expected calibration error by group.}  Partition $[0,1]$ into $M$
bins $\{B_m\}_{m=1}^M$.  For group $a$:
\begin{equation}
  \ece_a = \sum_{m=1}^{M}
    \frac{|\{i\in\mathcal{D}_a:\hat{p}_i\in B_m\}|}{|\mathcal{D}_a|}
    \bigl|\bar{y}(B_m,a)-\bar{p}(B_m,a)\bigr|,
  \label{eq:ece}
\end{equation}
where $\bar{y}(B_m,a)$ and $\bar{p}(B_m,a)$ are the mean outcome and mean
predicted probability in bin $m$ restricted to group $a$.  We compute
\eqref{eq:ece} under both equal-width and equal-mass (quantile) bins and
report the agreement, because the binning scheme is a known source of ECE
bias~\cite{nixon2019}.

\textit{4) Maximum calibration error by group.}  A bin-averaged statistic can
conceal a single badly miscalibrated region, so we also report
\begin{equation}
  \mce_a = \max_{m:\,|B_m\cap\mathcal{D}_a|>0}
    \bigl|\bar{y}(B_m,a)-\bar{p}(B_m,a)\bigr|.
  \label{eq:mce}
\end{equation}
\S\ref{sec:usecases_results} identifies a decision context in which
\eqref{eq:mce} carries the decision weight and \eqref{eq:ece} does not.

\subsection{Audit questions}

The audit does not optimize a model.  Given a fixed deployed model $f_\theta$,
it asks five questions, each answered by a designated section.

\begin{enumerate}[(Q1)]
\item Which groups exhibit statistically significant disparities, on which
      metric, and what governs whether a real disparity is detectable at all?
      (\S\ref{sec:baseline}, \S\ref{sec:evidence})
\item Are the calibration conclusions stable under the analyst's free choices,
      chiefly the binning scheme? (\S\ref{sec:calibration})
\item Does a post-hoc mitigation reduce held-out EOD in \emph{every} run under
      a prespecified accuracy constraint, and with what uncertainty?
      (\S\ref{sec:mitigation})
\item Is the observed pattern more consistent with proxy sensitivity or excess
      label disagreement, and does that verdict survive violation of its
      assumptions? (\S\ref{sec:attribution})
\item Is there evidence of longitudinal drift in group-stratified performance,
      and at what false-alarm cost? (\S\ref{sec:monitoring})
\end{enumerate}

% ============================================================
%  IV. CLINICAL DECISION CONTEXT
% ============================================================
\section{Which Metric the Decision Depends On}
\label{sec:clinical_context}

A subgroup fairness audit is actionable only if the metric it prioritizes
matches the decision the score actually drives.  We therefore characterize
three representative deployed clinical prediction tasks along six dimensions:
who reads the score, when it is produced, what action it triggers, the harm of
a false negative, the harm of a false positive, and the subgroup metric whose
degradation would be most consequential.  The asymmetry between the two error
costs is what selects the metric, and that asymmetry differs across the three
tasks.

\paragraph{30-day hospital readmission}
\textit{Model user:} Discharge planners, case managers, and post-acute care
coordinators, who use the risk score to decide how much follow-up support a
patient needs before leaving the hospital. \\
\textit{Prediction timing:} During the index admission, in the final days
before discharge, once comorbidities, SDOH, the post-discharge plan, and
laboratory trends are available in the EHR. \\
\textit{Action after alert:} A high-risk score triggers automated care
pathways: accelerated primary-care follow-up, home-health nursing visits,
medication management, and enrollment in remote monitoring. \\
\textit{False-negative harm:} A genuinely high-risk patient is sent home
without extra support, so an unrecognized secondary infection or medication
problem can continue unchecked, leading to unplanned readmission or a
preventable death. \\
\textit{False-positive harm:} Low-risk patients consume limited home-health
and case-management capacity, coordinators experience workload fatigue, and
patients face unnecessary follow-up burden. \\
\textit{Priority metric and why:} Missed-case rate, evaluated as the equalized
false-negative rate.  The SDOH variables that drive these models, such as
housing instability, transportation access, and insurance status, are already
correlated with under-resourced groups.  A model that under-flags high-risk
patients in one subgroup denies that group the intervention intended to close
the gap.  It worsens the existing disparity, where a false positive only
wastes resources~\cite{obermeyer2019}.  Published readmission models are
predominantly discriminative rather than calibrated for resource targeting,
and their reported performance varies substantially across
populations~\cite{kansagara2011}.

\paragraph{Sepsis early warning}
\textit{Model user:} Bedside nurses, rapid-response teams, and emergency
department or ICU physicians, who receive real-time alerts prompting immediate
bedside evaluation. \\
\textit{Prediction timing:} Continuously throughout an ED or ICU stay as new
vital signs, laboratory values, and notes enter the EHR, in principle before a
clinician would independently suspect sepsis, defined per the Sepsis-3
consensus criteria as life-threatening organ dysfunction caused by a
dysregulated host response to infection~\cite{seymour2016,kipnis2016}. \\
\textit{Action after alert:} Sepsis screening orders, lactate draws, blood
cultures, and rapid-response activation to begin antibiotics and fluids
without delay. \\
\textit{False-negative harm:} A true sepsis case goes undetected.  Each
additional hour of delay from hypotension onset to effective antimicrobial
therapy is associated with a mean decrease in survival of roughly 7.6\% in
septic shock~\cite{kumar2006}, and shorter time to antibiotic administration is
associated with lower risk-adjusted in-hospital mortality in mandated sepsis
care~\cite{seymour2017}, so a missed or late alert can be directly
life-threatening. \\
\textit{False-positive harm:} Unnecessary antibiotic orders, excessive blood
draws, and false alarms that produce alert fatigue, which can lead clinicians
to ignore or override subsequent genuine alerts. \\
\textit{Priority metric and why:} Missed-case rate, at a fixed alarm rate.
Given how sharply mortality risk rises with each hour of delay, an
early-warning tool that under-detects sepsis in a particular subgroup puts
that group at a time-critical mortality risk, which a difference in
false-alarm rate would not.  Neonates are the clearest case, since their
baseline vital signs differ sharply from the adult population these models are
mostly trained on; the same argument covers any group underrepresented in a
model's training data.  Sepsis prediction is
also the setting where such models are furthest along toward deployment, and
reviews of the area single out validation gaps and algorithmic bias as the
open problems~\cite{haas2022,papareddy2025}.  One proprietary sepsis model is among the most widely deployed clinical
prediction tools in US hospitals, which is what makes its independent
evaluation so useful: external validation found substantially poorer
discrimination than its developer had reported, with a large fraction of
sepsis cases missed at the recommended alerting threshold~\cite{wong2021}.
That is the failure mode a stratified sensitivity audit is designed to
surface.

\paragraph{In-hospital mortality prediction}
\textit{Model user:} ICU attendings and critical-care teams, who use the
continuous risk score to inform resource allocation, escalation of care, and
goals-of-care conversations with families (the discussions that set the
objectives guiding treatment). \\
\textit{Prediction timing:} Throughout the ICU stay, updated continuously and
in some models hourly, with reported lead times ranging from roughly 6 to 48
hours before severe deterioration~\cite{escobar2020}. \\
\textit{Action after alert:} Pre-emptive ICU transfer, closer monitoring,
earlier initiation of vasopressors or renal replacement therapy, and earlier
goals-of-care discussions with the patient's family. \\
\textit{False-negative harm:} A deteriorating patient is left at a lower level
of care, so decline is recognized later, possibly past the point at which
escalation still changes the outcome. \\
\textit{False-positive harm:} A patient who would have survived is
over-prioritized, potentially prompting premature or overly intense
goals-of-care conversations, unnecessary ICU bed occupancy, or more invasive
treatment than warranted.  Treatment intensity at the end of life is not
race-neutral even absent any algorithmic input, so a subgroup-differential
false-positive rate would compound a pre-existing disparity rather than
introduce a new one~\cite{barnato2007}. \\
\textit{Priority metric and why:} Calibration.  Unlike a binary alert, this
score is consumed as a continuous probability that informs resource allocation
and end-of-life conversations, so the numeric value itself carries decision
weight beyond the implied ranking.  Established severity scores such
as SOFA~\cite{vincent1996} and APACHE~II~\cite{knaus1985} show limited
generalization across cohorts, and the prognostic accuracy of severity scores
for in-hospital mortality varies by the population in which they are
applied~\cite{raith2017}; a systematic review of AI mortality and outcome
models in ICU sepsis reports the same pattern of limited external
validation~\cite{mazza2026}.  A model whose predicted probabilities are
systematically too high or too low for a specific subgroup will misinform
these probability-dependent decisions even when its rank-ordering, and
therefore its AUC, remains acceptable~\cite{sjoding2020}.

% ============================================================
%  V. AUDIT FRAMEWORK
% ============================================================
\section{Audit Framework}
\label{sec:framework}

\subsection{Overview}

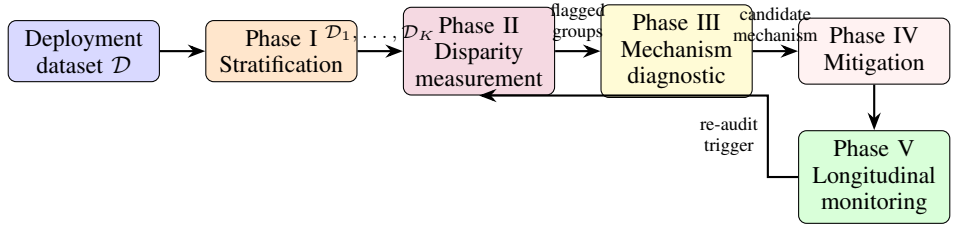
\begin{figure*}[t]
\centering
\begin{tikzpicture}[
  box/.style={draw, rounded corners=3pt, minimum width=2.0cm, minimum height=0.7cm,
              align=center, font=\small},
  pipeblue/.style={box, fill=blue!15},
  pipeorange/.style={box, fill=orange!20},
  pipepurple/.style={box, fill=purple!15},
  pipegreen/.style={box, fill=green!15},
  pipeyellow/.style={box, fill=yellow!20},
  pipepink/.style={box, fill=pink!20},
  arr/.style={-Stealth, thick},
  node distance=0.45cm and 0.6cm
]
\node[pipeblue]   (data)  {Deployment\\dataset $\mathcal{D}$};
\node[pipeorange, right=of data]  (p1) {Phase I\\Stratification};
\node[pipepurple, right=of p1]    (p2) {Phase II\\Disparity\\measurement};
\node[pipeyellow, right=of p2]    (p3) {Phase III\\Mechanism\\diagnostic};
\node[pipepink,   right=of p3]    (p4) {Phase IV\\Mitigation};
\node[pipegreen,  below=0.6cm of p4] (p5) {Phase V\\Longitudinal\\monitoring};

\draw[arr] (data) -- (p1);
\draw[arr] (p1)   -- node[above,font=\scriptsize,align=center]{$\mathcal{D}_1,\ldots,\mathcal{D}_K$} (p2);
\draw[arr] (p2)   -- node[above,font=\scriptsize,align=center]{flagged\\groups} (p3);
\draw[arr] (p3)   -- node[above,font=\scriptsize,align=center]{candidate\\mechanism} (p4);
\draw[arr] (p4)   -- (p5);
\draw[arr] (p5.west) -- ++(-0.4,0) |- node[near start, left, font=\scriptsize, align=center]
           {re-audit\\trigger} (p2.south);
\end{tikzpicture}
\caption{Five-phase audit pipeline.  Phases I--IV are applied at deployment
  time.  Phase V runs continuously post-deployment and triggers re-entry into
  Phase II when a CUSUM alarm fires.}
\label{fig:pipeline}
\end{figure*}

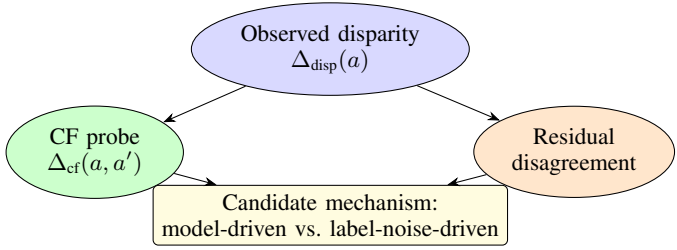
\begin{figure}[t]
\centering
\resizebox{\columnwidth}{!}{%
\begin{tikzpicture}[
  node distance=0.5cm and 1.0cm,
  oval/.style={draw, ellipse, minimum width=2.4cm, minimum height=0.7cm,
               align=center, font=\small},
  rect/.style={draw, rounded corners=2pt, minimum width=2.8cm,
               minimum height=0.7cm, align=center, font=\small},
  arr/.style={-Stealth}
]
\node[oval, fill=blue!15]   (obs) {Observed disparity\\$\Delta_{\text{disp}}(a)$};
\node[oval, fill=green!20, below left=of obs]  (cf)   {CF probe\\$\Delta_{\text{cf}}(a,a')$};
\node[oval, fill=orange!20, below right=of obs] (lne) {Residual\\disagreement};
\node[rect, fill=yellow!15, below=1.2cm of obs] (dec) {Candidate mechanism:\\model-driven vs.\ label-noise-driven};
\draw[arr] (obs) -- (cf);
\draw[arr] (obs) -- (lne);
\draw[arr] (cf)  -- (dec);
\draw[arr] (lne) -- (dec);
\end{tikzpicture}%
}
\caption{Disparity decomposition in Phase III\@.  Both probes are evaluated
  relative to a matched no-disparity control.}
\label{fig:attribution}
\end{figure}

The pipeline has five phases.  Fig.~\ref{fig:pipeline} shows the execution
flow and Fig.~\ref{fig:attribution} the disparity decomposition within
Phase~III.

\subsection{Phase I: Subgroup stratification}

Phase I partitions observations by sensitive attribute.  Groups smaller than a
minimum support threshold $n_{\min}$ are flagged instead of tested, and we
report a simulation-based power curve for detecting a prespecified difference
in true-positive rates.  This support check is descriptive: it does not turn
an underpowered estimate into evidence that no disparity exists.
\S\ref{sec:support} makes the check quantitative by pairing each axis with its
own minimum detectable effect.

\subsection{Phase II: Multi-metric disparity measurement}

For each group, Phase II computes subgroup AUC, ECE, MCE, TPR, and FPR\@.  The
maximum pairwise EOD is the prespecified inferential endpoint.
\begin{itemize}
\item \textbf{AUC disparity:} Wilcoxon--Mann--Whitney rank estimator, reported
      descriptively by group and as the maximum pairwise gap.
\item \textbf{EOD:} one-sided conditional permutation test
      ($n_{\text{perm}}=5{,}000$) shuffling group labels within
      outcome-by-seed strata, with Benjamini-Hochberg FDR
      correction~\cite{bh1995} at $\alpha=0.05$ across the prespecified HGBT
      disease-axis comparisons.
\item \textbf{Calibration:} equal-width and equal-mass ECE with $M=10$, MCE,
      and reliability curves, all reported descriptively.
\end{itemize}
The confirmatory flag is the EOD permutation test after FDR correction; AUC
gaps and calibration provide complementary descriptive evidence.

\subsection{Phase III: Mechanism diagnostics}

For each flagged group, Phase III applies two probes and classifies the
setting relative to a matched no-disparity control with the same disease and
seed.  The control is what prevents legitimate risk gradients, which also
activate the counterfactual probe, from being read as injected disparity.

\textit{1) Counterfactual proxy probe.}  For each observation
$(\mathbf{x}_i,a_i)$ with $a_i=a$, we measure the mean prediction change when
proxy features associated with the disadvantaged group are varied while labels
are held fixed:
\begin{equation}
  \Delta_{\text{cf}}(a,a')=\frac{1}{|\mathcal{D}_a|}
    \sum_{i\in\mathcal{D}_a}\bigl|f_\theta(\tilde{\mathbf{x}}_i)-f_\theta(\mathbf{x}_i)\bigr|,
  \label{eq:cf}
\end{equation}
where $\tilde{\mathbf{x}}_i$ substitutes the proxy columns of the
disadvantaged group with reference-group values.  The excess probe value
$\Delta_{\text{cf}}^{\text{excess}}=\Delta_{\text{cf}}-\Delta_{\text{cf}}^{\text{control}}$
is taken relative to the matched control.

\textit{2) Residual-disagreement probe.}  Defined as the target group's mean
$|y-\hat{p}|$ minus the corresponding mean outside the target group, with an
analogous excess
$\hat{\eta}^{\text{excess}}=\hat{\eta}-\hat{\eta}^{\text{control}}$.  If
$\Delta_{\text{cf}}^{\text{excess}}\ge\hat{\eta}^{\text{excess}}$ at a fixed
margin $\tau=0.03$, the setting is labeled model-driven; otherwise
label-noise-driven.  When both excesses are near zero the label is
no-material-disparity.

The rule returns exactly one label.  \S\ref{sec:attribution} shows that this
is its central weakness.

\subsection{Phase IV: Post-hoc mitigation}
\label{sec:phase4}

Phase IV applies two post-hoc procedures in sequence, neither of which
retrains the base model.

\textit{1) Weighted calibration.}  We fit a propensity model
$e(\mathbf{x})=\mathbb{P}(A=a\mid\mathbf{x})$ by logistic regression on the
calibration set, and weight calibration observations by
\begin{equation}
  w_i=\frac{\tilde{\pi}_a}{\hat{e}(\mathbf{x}_i)},
  \label{eq:ipw}
\end{equation}
where $\tilde{\pi}_a$ is the marginal prevalence of group
$a$~\cite{kamiran2012}.

\textit{2) Group-wise Platt scaling.}  For each group $a$ we fit
\begin{equation}
  \tilde{p}_i=\sigma\!\bigl(\alpha_a\cdot\text{logit}(\hat{p}_i)+\beta_a\bigr)
  \label{eq:platt}
\end{equation}
on the IPW-weighted calibration set~\cite{guo2017}.  This adjusts calibration
per group while preserving rank order within the group, and hence within-group
AUC.

\textit{3) Threshold optimization.}  We solve for group-specific thresholds
$\{\tau_a\}$ minimizing EOD subject to overall accuracy remaining at or above
96\% of uncorrected accuracy, following Hardt et al.~\cite{hardt2016}.  The
search is coordinate descent over a 97-point grid on $[0.02,0.98]$ with four
sweeps, which is deterministic but returns a local rather than a global
optimum, so the reported reductions are lower bounds on what the constrained
objective admits.  Because the two procedures are applied in sequence,
threshold optimization inherits the Platt calibration and the two
configurations share a calibration result.

\subsection{Phase V: Longitudinal drift monitoring}

Phase V tracks group-stratified equalized-odds streams over rolling windows of
$W=1{,}200$ patients.  For a monitored statistic $S_t$, the EOD between the
composite Medicaid/uninsured group and commercially insured patients, the
one-sided CUSUM detector maintains
\begin{equation}
  C_t=\max\!\bigl(0,\;C_{t-1}+(S_t-\mu_0)-\kappa\bigr),
  \label{eq:cusum}
\end{equation}
where $\mu_0$ is the in-control mean estimated over 8 baseline windows and
$\kappa=0.005$ is the allowance parameter.  An alarm fires when $C_t\ge h$,
with $h=\max(0.15,\,c\cdot\hat{\sigma}_0)$ set adaptively from the in-control
standard deviation~\cite{page1954}.  The multiplier $c$ is the monitor's
single free parameter and is swept in \S\ref{sec:monitoring}.  Separate
streams run per disease.  The injected scenario at window $t^*=20$
progressively suppresses true-positive predictions for the disadvantaged group
by 0.9 score units.  EOD widens as a result, and catching that widening is
what Phase V is for.

Two conventional properties characterize such a monitor: detection delay after
a real change, and in-control average run length $\text{ARL}_0$, the expected
number of windows between false alarms when nothing has
changed~\cite{montgomery2009}.  We report both, because either alone permits a
misleading operating point.

% ============================================================
%  VI. EXPERIMENTAL SETUP
% ============================================================
\section{Experimental Setup}

All main claims come from the synthetic SDOH benchmark described below.  Code,
configuration files, and seed-level logs are available at~\cite{kaiser_repo};
see \S\ref{sec:reproducibility}.

\subsection{Synthetic SDOH benchmark}

\begin{figure}[t]
\centering
\includegraphics[width=0.9\columnwidth]{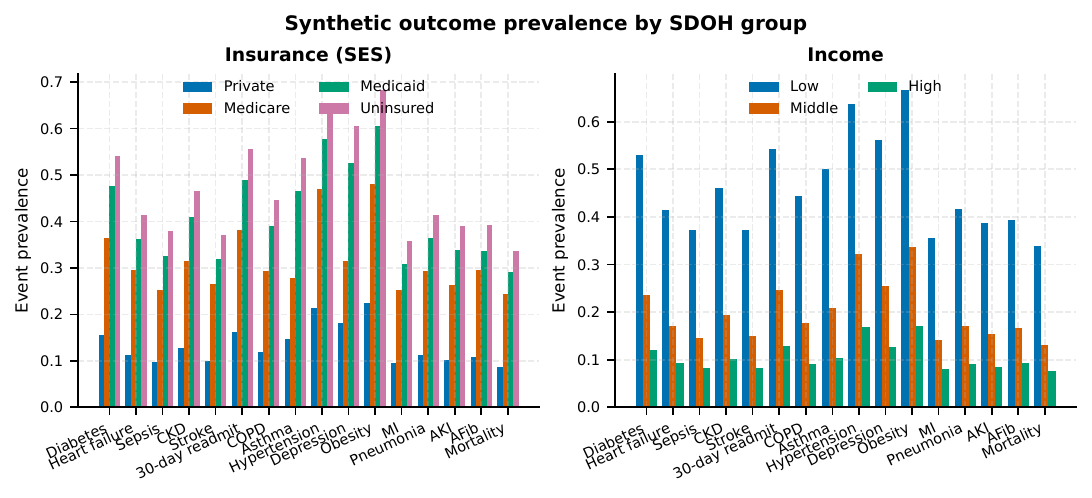}
\caption{Outcome prevalence by SDOH axis across all 16 disease tasks (HGBT,
  model-driven setting, 3 seeds).  Prevalence ratios between the most- and
  least-disadvantaged groups range from $1.5\times$ to $2.5\times$, which is
  the latent SES gradient built into the data-generating process.}
\label{fig:prevalence}
\end{figure}

We generate cohorts of $N=12{,}000$ patients per disease from a structural
causal model with correlated SDOH attributes.  A latent socioeconomic (SES)
factor is drawn per patient, anchored on a race-conditional mean to encode
structural inequity.  Income, education, and area deprivation are derived from
the SES factor, and insurance type is derived from age (Medicare eligibility)
and SES (Medicaid or uninsured at low SES), so that the 15 axes covary.  The
binary outcome is generated as
\begin{equation}
  Y_i\sim\text{Bernoulli}\!\bigl(\sigma\bigl(\mathbf{c}^\top\mathbf{x}_i^{\text{clin}}
    +\lambda\cdot\bar{g}(a_i)+\varepsilon_i\bigr)\bigr),
  \label{eq:dgp}
\end{equation}
where $\mathbf{c}$ is a shared clinical coefficient vector, $\bar{g}(a_i)$ is
a mean-centered SDOH gradient scaled by $\lambda=0.6$ to keep within-disease
prevalence ratios in a clinically plausible 1.5--2.5$\times$ range, and
$\varepsilon_i$ is individual noise.  Disease-specific base log-odds set
overall prevalences in the range 0.17--0.38.  The deployment feature matrix
excludes all 15 sensitive attributes; disparity arises through
group-correlated clinical proxy columns.  Fig.~\ref{fig:prevalence} shows the
resulting outcome-prevalence gradient across axes.

\textit{Injected disparity settings.}  Three settings are evaluated per
disease: (i)~\emph{none}, with gradients only and equal low label noise
($\eta\approx0.04$), which supplies the matched control; (ii)~\emph{model-driven},
where proxy columns are inflated for the disadvantaged group after labels are
fixed, so the model over-relies on a misleading proxy while true risk is
unchanged; (iii)~\emph{label-noise}, with an elevated label-flip rate
($\eta\approx0.40$) for uninsured patients, modeling worse documentation
quality.  Phase III should distinguish (ii) from (iii);
\S\ref{sec:attribution} tests whether it does, and what happens when both are
present at once.

\subsection{Baseline models}

We audit three model classes spanning common clinical AI deployments.
\begin{itemize}
\item \textbf{Logistic regression (LR):} $\ell_2$-penalized, no
      sensitive-attribute input.
\item \textbf{Histogram gradient boosted trees (HGBT):} scikit-learn
      \texttt{HistGradientBoostingClassifier}~\cite{sklearn2011} with native
      missing-value support.  HGBT is the designated model for all
      confirmatory inference.
\item \textbf{Neural network (MLP):} a dependency-light one-hidden-layer NumPy
      perceptron with 48 ReLU units, $\ell_2$ regularization, deterministic
      initialization, and full-batch gradient descent.
\end{itemize}
All three exclude the 15 sensitive attributes as direct inputs.  Disparity
still arises through group-correlated proxy features.  Dropping a protected
attribute is a well-known non-solution~\cite{obermeyer2019}.

\subsection{Statistical inference}

Held-out predictions from three independently generated seed cohorts are
pooled for confirmatory inference.  EOD 95\% confidence intervals come from a
non-parametric bootstrap ($n_{\text{boot}}=5{,}000$)~\cite{efron1994}.
Significance is assessed by one-sided conditional permutation tests
($n_{\text{perm}}=5{,}000$)~\cite{good2005} with Benjamini-Hochberg FDR
correction at $\alpha=0.05$~\cite{bh1995} across the prespecified HGBT
disease-axis comparisons.  With 5{,}000 permutations the smallest attainable
raw $p$-value is $2\times10^{-4}$, which is the resolution floor reported for
the socioeconomic axes.  Point estimates are mean $\pm$ sample standard
deviation over three seeds unless stated otherwise.

Mitigation effects are analyzed as \emph{paired} per-run differences rather
than as differences of independently reported configuration means, since the
same held-out cohort is scored under both configurations.  We report bootstrap
95\% intervals on the paired mean, Wilcoxon signed-rank $p$-values across the
16 disease-level deltas, and Clopper-Pearson intervals on the count of runs
improved.

\subsection{Subgroup support and detectable effects}
\label{sec:support}

\begin{figure}[t]
\centering
\includegraphics[width=0.9\columnwidth]{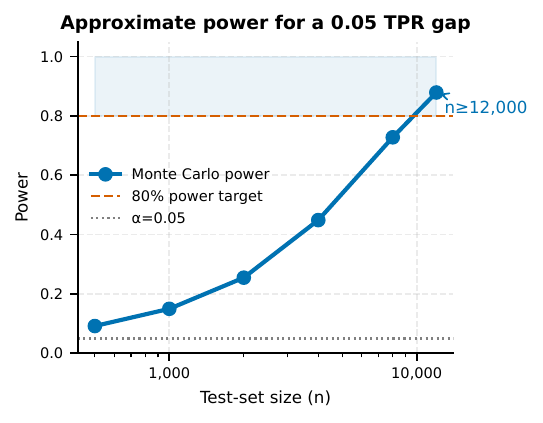}
\caption{Monte Carlo power curve for detecting a true EOD of 0.05 at
  $\alpha{=}0.05$.  The dashed line marks 80\% power; approximately
  $n{=}8{,}000$ per group is required.}
\label{fig:power_curve}
\end{figure}

Table~\ref{tab:power} reports Monte Carlo power for detecting an EOD of 0.05
at $\alpha=0.05$ as a function of per-group sample size, using the pooled
two-proportion z-test as a proxy.  Achieving 80\% power at that effect size
requires roughly $n=8{,}000$ per group (Fig.~\ref{fig:power_curve}).

That reference number sets up the point of this subsection.  In the benchmark
as run, \emph{no} subgroup cell reaches it: all 78 cells across the 18 axes
fall below $n=8{,}000$, the median cell holds 308 held-out patients, 54 cells
hold fewer than 500, and the smallest intersectional cell (Black \& Uninsured)
holds 30.  A single global power reference is therefore not the right
instrument.  What matters for each axis is its \emph{own} minimum detectable
effect, computed from the positives in its smallest cell, because the TPR term
of \eqref{eq:eod} is estimated on positives only.  We define for each axis
\begin{equation}
  \text{MDE} = (z_{1-\alpha/2}+z_{1-\beta})\sqrt{\tfrac{2p(1-p)}{n^{+}_{\min}}},
  \qquad
  R = \frac{\eod_{\max}}{\text{MDE}},
  \label{eq:mde}
\end{equation}
with $p=0.30$ the reference positive rate, $\beta=0.20$, and $n^{+}_{\min}$
the positive count in the axis's smallest cell.  Table~\ref{tab:support}
reports both.  $R$ is the axis-standardized effect size, and
\S\ref{sec:evidence} shows that it predicts which axes reach significance far
better than raw EOD does.

\begin{table}[t]
\caption{Monte Carlo power for detecting a true EOD of 0.05 at $\alpha=0.05$
  as a function of per-group sample size (10,000 simulation runs).  MDE is the
  minimum detectable EOD under the z-test approximation.  No subgroup cell in
  the benchmark reaches the $n=8{,}000$ needed for 80\% power at this effect
  size; see Table~\ref{tab:support}.}
\label{tab:power}
\centering\footnotesize
\setlength{\tabcolsep}{4pt}
\begin{tabular}{rcc}
\toprule
$n$ per group & Power & MDE \\
\midrule
500   & 0.09 & 0.065 \\
1,000 & 0.15 & 0.046 \\
2,000 & 0.26 & 0.032 \\
4,000 & 0.45 & 0.023 \\
8,000 & 0.73 & 0.016 \\
12,000 & 0.88 & 0.013 \\
\bottomrule
\end{tabular}
\end{table}

\subsection{Mitigation and monitoring protocol}

Phase IV uses a 70/15/15 train/calibration/test split.  IPW weights are
estimated on the calibration set, Platt parameters are fit on the weighted
calibration set, and evaluation is on the held-out test set.  Phase V uses
$W=1{,}200$, 8 in-control baseline windows, 40 windows per run, and slack
$\kappa=0.005$.  The adaptive threshold at the $6\sigma$ reference setting
averages $h=0.31$ across runs (range 0.19--0.40); the $4\sigma$ and $8\sigma$
settings average 0.21 and 0.41.

% ============================================================
%  VII. RESULTS
% ============================================================
\section{Results and Analysis}

Results address Q1--Q5 from \S\ref{sec:formulation}.  Unless noted, numbers
are mean $\pm$ standard deviation over three seeds on the held-out test set,
for the HGBT model and the insurance axis.

\subsection{Baseline disparities across SDOH axes}
\label{sec:baseline}

\begin{figure}[t]
\centering
\includegraphics[width=0.9\columnwidth]{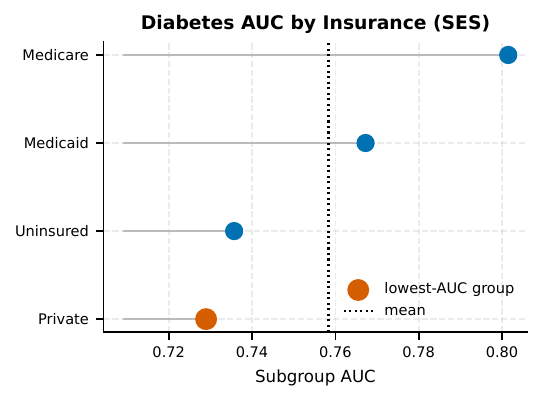}
\caption{Per-group AUC for the lead disease and axis (diabetes, insurance,
  HGBT, 3 seeds).  AUC gaps are smaller than EOD gaps because ranking
  performance is less sensitive to group-conditional threshold miscalibration
  than error-rate parity.}
\label{fig:group_auc}
\end{figure}

Table~\ref{tab:baseline} reports AUC and disparity metrics for all three model
classes under the model-driven setting on the diabetes task.  Across single
axes, averaged over 16 diseases, socioeconomic and material-hardship axes
produce the largest gaps: area deprivation index ($\eod=0.542\pm0.093$, peak
$0.682$ on depression), income ($0.511\pm0.082$), and insurance
($0.509\pm0.060$).  Biological sex shows substantially smaller disparity
($0.034\pm0.010$) and is the smallest axis in 15 of 16 diseases.  Per-group
AUC gaps are smaller than EOD gaps because ranking is less sensitive to
group-conditional threshold miscalibration (Fig.~\ref{fig:group_auc}).

\begin{figure}[t]
\centering
\includegraphics[width=0.9\columnwidth]{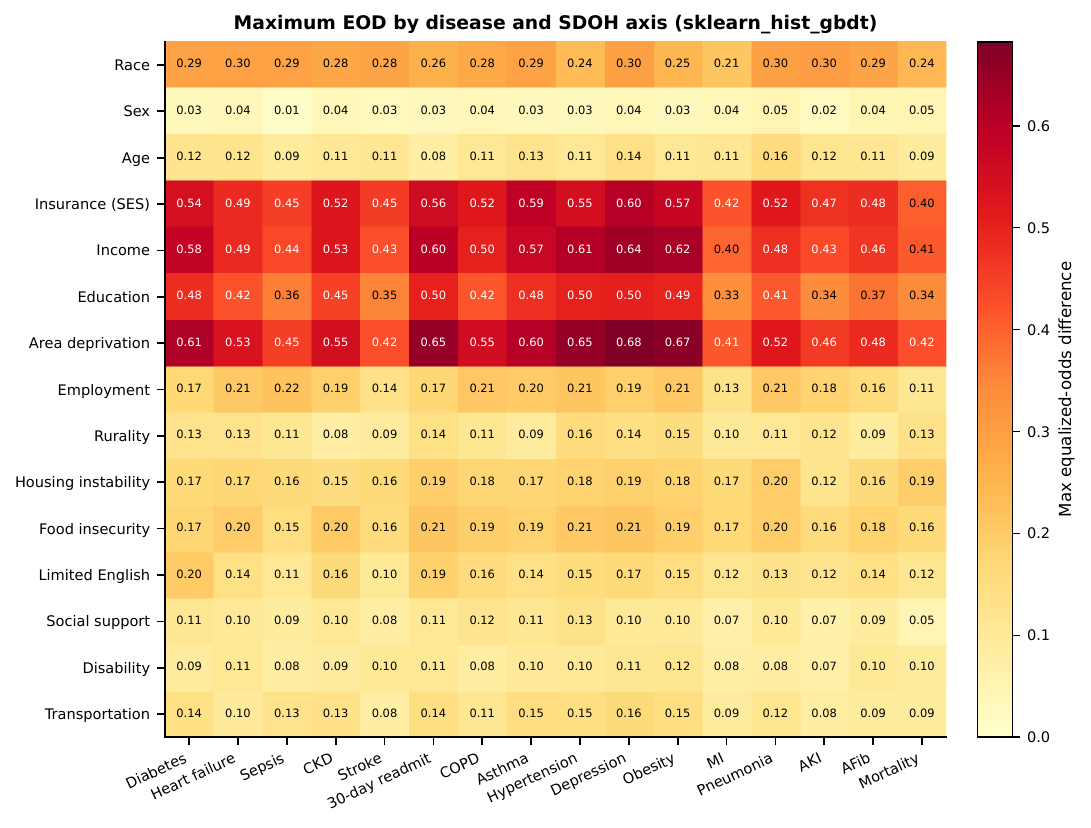}
\caption{Maximum pairwise equalized-odds difference across 16 diseases and the
  15 single SDOH axes for the HGBT model, averaged over three seeds.  The
  socioeconomic block is uniformly at the top, but its internal ordering
  varies by disease.}
\label{fig:heatmap}
\end{figure}

\begin{figure}[t]
\centering
\includegraphics[width=0.9\columnwidth]{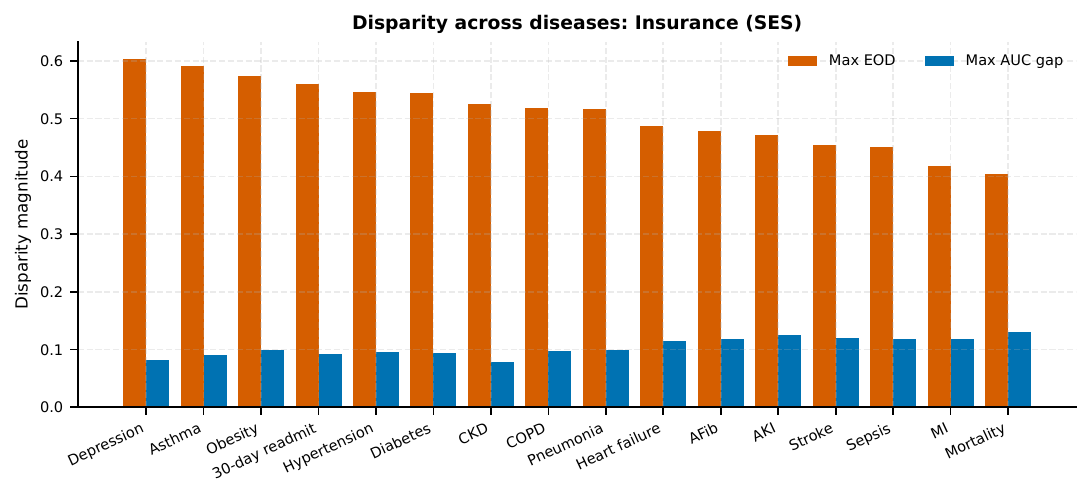}
\caption{Insurance-axis EOD across all 16 clinical tasks (HGBT, 3 seeds).
  Error bars show $\pm$1 SD.  The dashed line marks the cross-disease mean of
  $0.509$.}
\label{fig:disease_eod}
\end{figure}

The ranking is stable only in coarse terms.  The four socioeconomic axes (ADI,
income, insurance, education) occupy the top four positions in all 16 of 16
diseases, but their internal order is not fixed: the specific ordering
$\text{ADI}>\text{income}>\text{insurance}$ holds in only 7 of 16
(Fig.~\ref{fig:heatmap}).  Fig.~\ref{fig:disease_eod} gives the per-disease
spread on the insurance axis.  Claims are safe at the level of the axis block and unsafe at the level of an
individual axis rank.

\begin{table}[t]
\caption{Baseline fairness metrics across model classes, diabetes task,
  model-driven setting (3 seeds).  The first three rows are the insurance
  axis; the last row is the same HGBT model on the sex axis as a negative
  reference, where no disparity was injected.  $\Delta\text{AUC}$ is the
  largest pairwise AUC gap, $\eod$ the largest pairwise equalized-odds
  difference, and ECE the mean group ECE over the axis categories.  Overall
  AUC is a property of the model, so it is identical for the two HGBT rows.}
\label{tab:baseline}
\centering\footnotesize
\setlength{\tabcolsep}{3pt}
\begin{tabular}{llcccc}
\toprule
Model & Axis & AUC & $\Delta\text{AUC}$ & $\eod$ & ECE \\
\midrule
LR   & Insurance & $.816{\pm}.010$ & $.093{\pm}.016$ & $.603{\pm}.009$ & $.073{\pm}.008$ \\
MLP  & Insurance & $.791{\pm}.015$ & $.101{\pm}.019$ & $.554{\pm}.022$ & $.069{\pm}.013$ \\
HGBT & Insurance & $.804{\pm}.014$ & $.094{\pm}.030$ & $.544{\pm}.016$ & $.072{\pm}.006$ \\
\midrule
HGBT & Sex & $.804{\pm}.014$ & $.023{\pm}.027$ & $.032{\pm}.018$ & $.038{\pm}.007$ \\
\bottomrule
\end{tabular}
\end{table}

\subsection{What subgroup significance actually tracks}
\label{sec:evidence}

Disparity magnitude and statistical evidence rank the axes differently, and
the difference has a systematic cause.  Table~\ref{tab:support}
reports both orderings together with the support that produced them.

The largest gaps in the benchmark turn out to be the least evidenced.  The three prespecified intersections carry the largest raw
gaps of any axis, with race$\times$insurance averaging $0.703\pm0.031$ across
diseases, above every single axis, but our inference protocol prespecified
permutation testing for the 15 single axes only, so they carry no $p$-value.
This limitation belongs to the present run, not to the data, and it has a
reason: the intersections have 8 to 14 cells against the 2 to 5 of a typical
single axis, and their smallest cells hold 30 to 45 patients, so testing them
would have required a prespecified minimum cell count that we did not set.  We report them descriptively and do not claim them
as findings.

Among the 15 tested axes, raw effect size predicts significance only weakly.  The rank correlation between an axis's
EOD and the number of diseases in which it is significant is $\rho=0.56$
($p=0.029$).  Race illustrates the failure: at $\eod=0.274$ it is the fifth
largest axis, yet reaches significance in only 11 of 16 diseases, while food
security at $\eod=0.185$ reaches 16 of 16.

Standardizing by each axis's own detectable floor largely resolves this.
Replacing EOD with $R=\eod/\text{MDE}$ from \eqref{eq:mde} raises the
rank correlation to $\rho=0.78$ ($p=0.0005$).  The two axes significant in no
disease, age and sex, have the two smallest standardized effects in the table
($R=0.34$ and $R=0.23$), strictly below every axis that reaches significance
anywhere, whose minimum is $R=0.46$.  Race's anomaly resolves in the same
direction: its five-way split leaves only about nine positives in its smallest
cell, giving $R=0.46$, near the bottom of the tested set despite a mid-table
raw gap.

A subgroup audit reporting only effect sizes and one reporting only
$p$-values mislead in opposite directions.  Neither is interpretable without
the cell support behind it.  Age is the
clearest case: an $\eod$ of $0.113$ that is significant in zero of 16 diseases
is not evidence that the age axis is clean, because the axis's own detectable
floor sits at $0.329$.  The audit cannot distinguish absence of disparity from
absence of resolution, and should say so.  We accordingly report
$n^{+}_{\min}$, MDE, and $R$ alongside every axis-level disparity, and treat
any axis with $R<1$ as underpowered for its own observed effect, which in
this benchmark is 12 of the 15 tested axes.

\begin{table}[t]
\caption{Equalized-odds disparity, subgroup support, and evidence by axis
  (HGBT, mean over 16 diseases), ordered by effect size.  $k$ is the number of
  cells, $n^{+}_{\min}$ the mean positive count in the smallest cell, MDE the
  axis's own minimum detectable EOD from \eqref{eq:mde}, and
  $R=\eod/\text{MDE}$ the standardized effect.  ``Sig.'' counts diseases
  significant after BH correction at $\alpha{=}0.05$.  Intersections were
  outside the prespecified inference protocol and are untested (n/a).  Effect
  order and evidence order differ; $R$ largely reconciles them
  (\S\ref{sec:evidence}).  Source:
  \texttt{outputs/sdoh\_multidisease/axis\_support.csv}.}
\label{tab:support}
\centering\footnotesize
\setlength{\tabcolsep}{3.5pt}
\begin{tabular}{lccccc}
\toprule
\textbf{Axis} & $\boldsymbol{\eod_{\max}}$ & $\boldsymbol{k}$ &
  $\boldsymbol{n^{+}_{\min}}$ & \textbf{MDE} & \textbf{Sig.} \\
\midrule
Race\,$\times$\,insurance & $0.703{\pm}0.031$ & 14 & 3 & 1.02 & n/a \\
Income\,$\times$\,race & $0.641{\pm}0.126$ & 12 & $<1$ & 2.98 & n/a \\
Insurance\,$\times$\,disability & $0.581{\pm}0.064$ & 8 & 9 & 0.62 & n/a \\
\midrule
Area deprivation & $0.542{\pm}0.093$ & 5 & 7 & 0.67 & 16/16 \\
Income & $0.511{\pm}0.082$ & 3 & 12 & 0.52 & 16/16 \\
Insurance & $0.509{\pm}0.060$ & 4 & 39 & 0.29 & 16/16 \\
Education & $0.422{\pm}0.064$ & 3 & 21 & 0.40 & 16/16 \\
Race & $0.274{\pm}0.027$ & 5 & 9 & 0.60 & 11/16 \\
Food security & $0.185{\pm}0.021$ & 2 & 123 & 0.16 & 16/16 \\
Employment & $0.181{\pm}0.033$ & 3 & 65 & 0.22 & 16/16 \\
Housing & $0.172{\pm}0.019$ & 2 & 73 & 0.21 & 16/16 \\
Language & $0.143{\pm}0.028$ & 2 & 80 & 0.20 & 16/16 \\
Transportation & $0.119{\pm}0.028$ & 2 & 97 & 0.18 & 16/16 \\
Rurality & $0.117{\pm}0.024$ & 3 & 76 & 0.21 & 11/16 \\
Age & $0.113{\pm}0.019$ & 4 & 30 & 0.33 & 0/16 \\
Disability & $0.095{\pm}0.014$ & 2 & 112 & 0.17 & 16/16 \\
Social support & $0.095{\pm}0.022$ & 2 & 121 & 0.17 & 12/16 \\
Sex & $0.034{\pm}0.010$ & 2 & 149 & 0.15 & 0/16 \\
\bottomrule
\end{tabular}
\end{table}

\subsection{Binning choice and worst-bin error}
\label{sec:calibration}

Equal-width ECE is sensitive to bin placement, so a calibration conclusion
that depends on it needs a robustness check~\cite{nixon2019}.  Across all
1{,}248 group-level cells we compute ECE under both equal-width and equal-mass
binning at $M=10$.  The two agree closely in ordering: cell-level rank
correlation is $\rho=0.96$, and the axis-level ordering they induce correlates
at $\rho=0.96$.  Magnitudes differ modestly and in a consistent direction,
with equal-mass ECE higher on average ($0.076$ against $0.069$; mean absolute
difference $0.010$, maximum $0.069$).  The most-miscalibrated third of the
axes is identical under both schemes, and no axis moves between the top and
bottom thirds; four axes exchange between the middle and bottom thirds, all of
them inside the narrow band $\ece\in[0.040,0.042]$ where the axes are
effectively tied.  The calibration conclusions in this paper are therefore not
an artifact of the binning choice, though absolute values should be read as
scheme-dependent to about $\pm0.01$.

Maximum calibration error behaves differently, and this matters because
\S\ref{sec:clinical_context} identifies a decision context, in-hospital
mortality, in which the worst bin carries the decision weight.  Mean MCE across cells is $0.341$, roughly five times
mean ECE: the worst-calibrated region of the probability range is badly wrong
even where the bin-averaged statistic looks acceptable.  MCE also ranks the
axes differently from ECE, correlating at only $\rho=0.65$ at the axis level.
An audit reporting ECE alone will understate the severity of the worst
miscalibration, and on some axes it will point at the wrong axis altogether.
We report both.  Table~\ref{tab:usecases} records which decision contexts
require which.

\subsection{Phase IV mitigation across all 48 runs}
\label{sec:mitigation}

\begin{figure}[t]
\centering
\includegraphics[width=0.9\columnwidth]{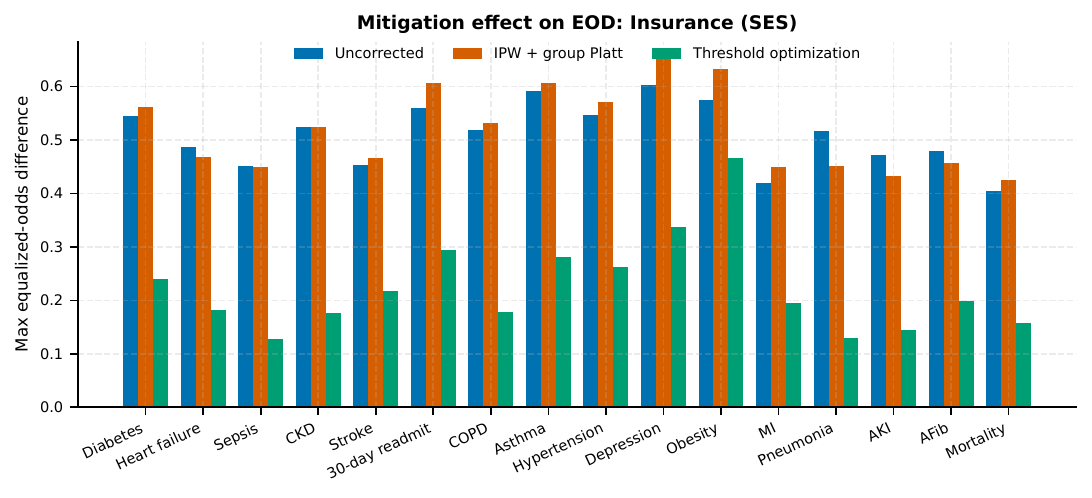}
\caption{Maximum equalized-odds difference before and after mitigation on the
  insurance axis.  Thresholds are selected on the calibration split and
  evaluated on the held-out test split.}
\label{fig:mitigation_eod}
\end{figure}

\begin{figure}[t]
\centering
\includegraphics[width=0.9\columnwidth]{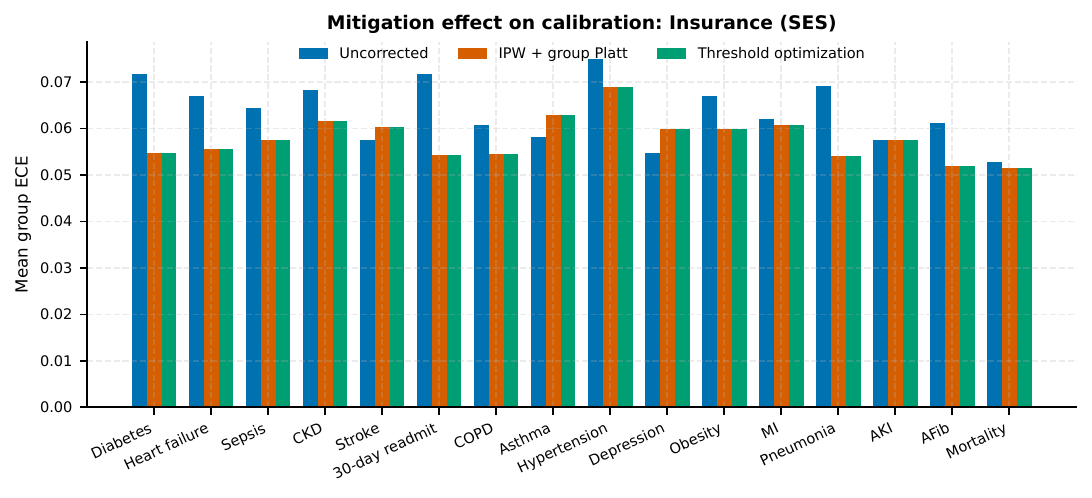}
\caption{Mean group expected calibration error for the same comparison.
  Group-wise Platt scaling targets calibration rather than EOD.}
\label{fig:mitigation_ece}
\end{figure}

\begin{figure}[t]
\centering
\includegraphics[width=0.9\columnwidth]{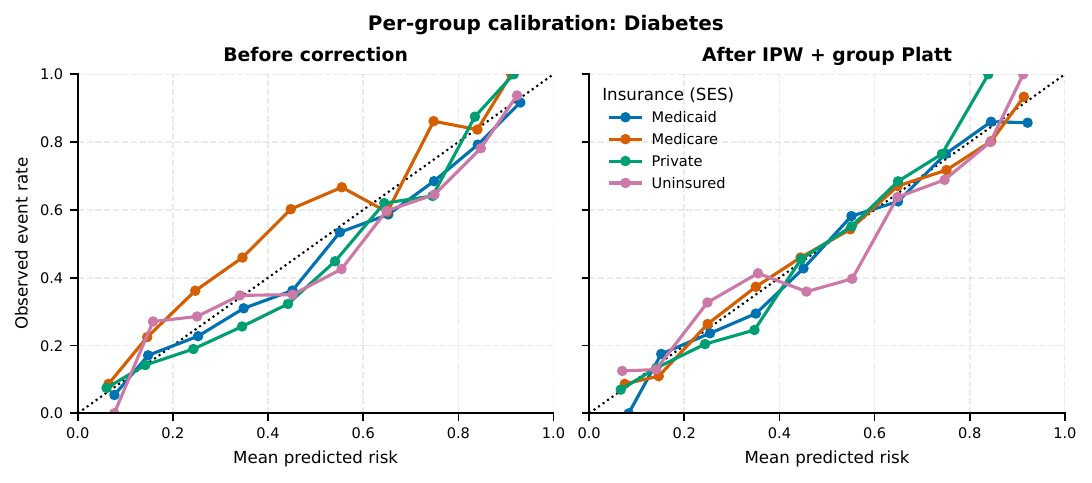}
\caption{Insurance-stratified reliability curves before and after IPW-weighted
  group Platt scaling for the lead disease.  Curves closer to the diagonal are
  better calibrated.}
\label{fig:calibration}
\end{figure}

Table~\ref{tab:mitigation} reports the two-step mitigation on the diabetes
task and Table~\ref{tab:mitigation_extended} across all 48 disease-seed runs, with
Figs.~\ref{fig:mitigation_eod}--\ref{fig:calibration} showing the same
comparison for EOD, ECE, and the reliability curves.  What matters is whether a method improves the target metric in \emph{every}
held-out run, since a deployment sees one run and never the mean.

Threshold optimization reduced EOD in 48 of 48 runs and in all 16 of 16
diseases, with a paired mean $\Delta\eod=-0.285$ (95\% CI $[-0.313,-0.252]$;
Wilcoxon $p=3.1\times10^{-5}$ across the 16 disease-level deltas), improvement
rate 48/48 (Clopper-Pearson 95\% CI $[0.93,1.00]$), and no cost in overall
AUC, which rose slightly ($+0.004$, CI $[+0.003,+0.005]$) through the Platt
stage it inherits.

Weighted group-wise Platt scaling behaved very differently, and a means-only
table would describe its failure too broadly.  It improved EOD in 19 of 48
runs (CI $[0.26,0.55]$) and worsened it in 28, with
one run unchanged; at the disease level it improved 6 of 16.  Its paired mean
effect was $+0.009$ with 95\% CI $[-0.007,+0.024]$ and Wilcoxon $p=0.30$.
That interval contains zero, so the data will not support a claim that Platt
scaling \emph{worsens} EOD on average.  They support a claim of
\emph{unreliability}: its improvement rate is statistically indistinguishable
from a coin flip, and its run-to-run spread on EOD (SD $0.033$ across
disease-level deltas, against a mean of $0.009$) runs to several times its
central effect.  Wherever the mean sits, a method whose 95\% interval spans
both meaningful improvement and meaningful harm makes a poor default.

The same method is nonetheless the better calibrator, improving ECE in 33 of
48 runs (CI $[0.54,0.81]$) and 13 of 16 diseases, with paired mean
$\Delta\ece=-0.006$ (CI $[-0.009,-0.002]$, $p=0.004$).

This divergence is the substantive finding, and it comes from a difference
between the two objectives; neither method is defective.  Platt scaling
recalibrates per-group probability magnitudes without constraining
class-conditional error rates, so a monotone rescaling that lowers ECE can
move a group's TPR and FPR in opposite directions and widen EOD.  Threshold
optimization operates directly on the EOD objective subject to the accuracy
constraint of \S\ref{sec:phase4}, and inherits whatever calibration gain the
Platt stage supplied, which is why the two configurations share an ECE column.
The audit consequence is concrete: a calibration improvement is no evidence of
a fairness improvement, and a report showing only ECE would have presented an
EOD-neutral-to-harmful intervention as a success.

\begin{table}[t]
\caption{Fairness metrics before and after Phase IV mitigation (HGBT,
  diabetes, insurance axis, 3 seeds).}
\label{tab:mitigation}
\centering\footnotesize
\setlength{\tabcolsep}{4pt}
\begin{tabular}{lccc}
\toprule
Configuration & AUC & EOD & ECE \\
\midrule
Uncorrected         & $.804{\pm}.014$ & $.544{\pm}.016$ & $.072{\pm}.006$ \\
IPW+group Platt     & $.808{\pm}.015$ & $.561{\pm}.072$ & $.055{\pm}.003$ \\
Threshold opt.      & $.808{\pm}.015$ & $.241{\pm}.079$ & $.055{\pm}.003$ \\
\bottomrule
\end{tabular}
\end{table}

\begin{table}[t]
\caption{Mitigation reliability and paired effects across 16 diseases on the
  insurance axis (HGBT, 3 seeds; 48 runs).  The $\Delta$ rows give the paired
  mean change with a bootstrap 95\% CI over the 16 disease-level deltas; lower
  EOD and ECE is better.  ``Impr.'' counts individual runs improved out of 48
  with a Clopper-Pearson 95\% CI\@.  The two configurations are applied
  sequentially, so threshold optimization inherits the Platt calibration and
  they share an ECE column.  The Platt EOD interval contains zero, so the
  defect to report is its variance.  Source:
  \texttt{outputs/sdoh\_multidisease/correction.csv}.}
\label{tab:mitigation_extended}
\centering\footnotesize
\setlength{\tabcolsep}{3pt}
\resizebox{\columnwidth}{!}{%
\begin{tabular}{lccccc}
\toprule
\textbf{Configuration} & \textbf{AUC} & \textbf{EOD} & \textbf{ECE} &
  \textbf{EOD impr.} & \textbf{ECE impr.} \\
\midrule
Uncorrected & $.796{\pm}.013$ & $.509{\pm}.060$ & $.064{\pm}.007$ & --- & --- \\
IPW\,+\,group Platt & $.799{\pm}.012$ & $.518{\pm}.078$ & $.058{\pm}.005$ &
  19/48 & 33/48 \\
\quad paired $\Delta$ & $+.004$ & $+.009$ &
  \textcolor{green!60!black}{$-.006$} & & \\
\quad 95\% CI & $[.003,.005]$ & \textcolor{red}{$[-.007,.024]$} &
  $[-.009,-.002]$ & $[.26,.55]$ & $[.54,.81]$ \\
Threshold opt. & $.799{\pm}.012$ & $.224{\pm}.089$ & $.058{\pm}.005$ &
  48/48 & 33/48 \\
\quad paired $\Delta$ & $+.004$ & \textcolor{green!60!black}{$-.285$} &
  \textcolor{green!60!black}{$-.006$} & & \\
\quad 95\% CI & $[.003,.005]$ & $[-.313,-.252]$ & $[-.009,-.002]$ &
  $[.93,1.00]$ & $[.54,.81]$ \\
\bottomrule
\end{tabular}%
}
\end{table}

\paragraph{Two-disease validation cohort}
The same comparison on the smaller validation run
(\texttt{outputs/sdoh\_validation/}; diabetes and sepsis, 3 seeds each) was
analyzed independently and reproduces the pattern.  For diabetes, EOD moved
from $0.602$ uncorrected to $0.681$ under Platt scaling (worse in 3 of 3
seeds) and to $0.347$ under threshold optimization (better in 3 of 3); for
sepsis, from $0.443$ to $0.411$ under Platt (better in 2 of 3) and to $0.157$
under threshold optimization (better in 3 of 3).  Pooling the six seeds,
threshold optimization improved EOD in 6 of 6 while Platt scaling worsened it
in 4 of 6.  Threshold optimization was therefore the method that most
consistently reduced EOD, reducing it in every seed across both diseases,
while Platt scaling targets probability calibration instead of equalized odds
and improved ECE in only 4 of 6 seeds.  By the rule that a method fails if it
worsens held-out runs, neither method succeeds on calibration, and only
threshold optimization succeeds on EOD.  The cautious
reading is that an improvement in one fairness measure does not imply an
improvement in another, so a result on one metric should not be generalized to
the others.  Agreement between a 6-run and a 48-run cohort on the direction of
both effects is reassuring, but neither licenses a deployment recommendation:
both draw on the same generator, and the accuracy constraint that makes
threshold optimization viable here is a property of that generator's group
sizes (\S\ref{sec:limitations}).

\subsection{Phase III mechanism diagnostics and their failure modes}
\label{sec:attribution}

\begin{figure}[t]
\centering
\includegraphics[width=0.9\columnwidth]{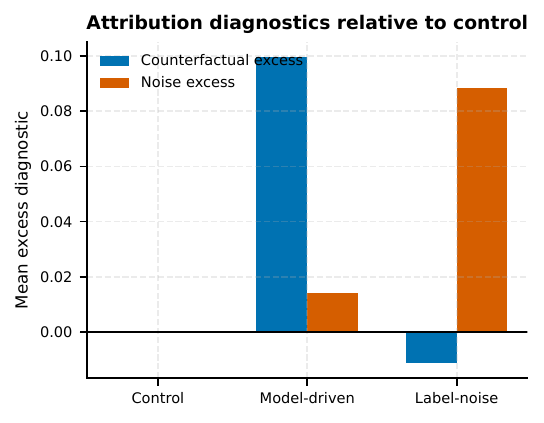}
\caption{Diagnostics relative to the matched control.  The model-driven
  condition raises the counterfactual-proxy probe, while the label-noise
  condition raises residual disagreement.}
\label{fig:attribution_bars}
\end{figure}

Under the controlled generator the diagnostic works.
Table~\ref{tab:attribution} reports the diabetes task: in the model-driven
setting $\Delta_{\text{cf}}$ rises from $0.282\pm0.012$ to $0.398\pm0.013$
while residual disagreement stays at the control level, and in the label-noise
setting the residual diagnostic rises from $0.104\pm0.014$ to
$0.187\pm0.021$ while counterfactual sensitivity does not.  The rule
classifies 144 of 144 disease-seed-setting cases correctly
(Fig.~\ref{fig:attribution_bars}).

That accuracy is a property of a generator in which exactly one mechanism is
active and the probe is handed the correct proxy columns.  Both assumptions
fail in practice, so we construct two conditions that violate them
deliberately (Table~\ref{tab:attribution_stress}).  Neither has a single
ground-truth label, so we report the label distribution in place of an
accuracy.

\begin{table}[t]
\caption{Phase III diagnostics under the controlled generator, diabetes task
  (3 seeds).  ``Correct'' means the diagnostic recovers the injected setting
  relative to the matched no-disparity control.}
\label{tab:attribution}
\centering\footnotesize
\begin{tabular}{lccc}
\toprule
Setting & $\Delta_{\text{cf}}$ & Est.\ $\hat{\eta}$ & Correct \\
\midrule
None (control)  & $0.282\pm0.012$ & $0.104\pm0.014$ & 100.0\% \\
Model-driven    & $0.398\pm0.013$ & $0.112\pm0.013$ & 100.0\% \\
Label-noise     & $0.262\pm0.003$ & $0.187\pm0.021$ & 100.0\% \\
\bottomrule
\end{tabular}
\end{table}

\begin{table}[t]
\caption{Stress tests across 16 diseases $\times$ 3 seeds (48 runs per
  condition).  \emph{Mixed mechanisms}: proxy sensitivity and label noise are
  both injected.  \emph{Misspecified proxy}: a model-driven disparity is
  present but the probe is given unrelated feature columns.  CF excess is the
  counterfactual probe magnitude net of its decision margin; a negative value
  means the probe registered less signal than the threshold for a material
  finding.  Source:
  \texttt{outputs/sdoh\_multidisease/attribution\_stress.csv}.}
\label{tab:attribution_stress}
\centering\footnotesize
\setlength{\tabcolsep}{4pt}
\begin{tabular}{p{2.35cm}cccc}
\toprule
\textbf{Condition} & \textbf{Model-} & \textbf{Label-} & \textbf{No mat.} &
  \textbf{CF} \\
 & \textbf{driven} & \textbf{noise} & \textbf{disparity} & \textbf{excess} \\
\midrule
Mixed mechanisms   & 21/48 & 27/48 & 0/48  & $+0.092$ \\
Misspecified proxy & 0/48  & 12/48 & 36/48 & $-0.235$ \\
\bottomrule
\end{tabular}
\end{table}

Under mixed mechanisms the diagnostic split almost evenly, 21 of 48
model-driven and 27 label-noise-driven, and never returned
no-material-disparity.  It correctly registered that something was wrong in
every run, but the split is an artifact of the single-label rule and not a
measurement of anything: both mechanisms were present in all 48 runs, so whichever label
is returned is at best half the answer, and the near-even division reflects
which mechanism happened to dominate a given seed.  There is no single correct
label when both mechanisms are active, and forcing one produces a false
binary; choosing one label when both are causing the problem splits the runs
between model and label noise without identifying a cause in either group.  A reader given only the
label would conclude that roughly half these cohorts had a label-quality
problem and the other half a model problem, when every one had both.

Under a misspecified proxy the failure is more consequential because it is
silent.  A model-driven disparity was present in all 48 runs, yet the probe
returned no-material-disparity in 36, label-noise-driven in 12, and
model-driven in none.  Mean CF excess fell from $+0.092$ to $-0.235$: probing
unrelated columns finds no sensitivity, and the rule reads absence of
sensitivity as absence of a model-driven problem.  The diagnostic depends on
correctly identifying the relevant proxy features, and without that it is
fragile: because it evaluates unrelated columns it cannot recognize the bias
that is in fact present, and the no-material-disparity verdicts are false
negatives dressed as clean results.  Nothing in the output
distinguishes ``we probed the right features and found nothing'' from ``we
probed the wrong features''.  The 12 runs relabeled as label-noise-driven are
worse still, because they redirect the investigation toward data quality when
the model is at fault.

Three consequences follow.  All three are limitations of the diagnostic; the
benchmark is doing its job.  The single-label output should be read as a
ranking of candidate explanations, since mixed-mechanism cohorts are common
in the field and the rule has no way to represent them.  Proxy selection
determines whether the probe is valid at all, so proxy columns have to be
prespecified, justified against domain knowledge before the audit runs, and
reported alongside the result; a no-material-disparity finding means nothing
without them.  And what the probe measures is the model's sensitivity to
feature values, which falls well short of establishing that group membership
caused the disparity through those features.  The perfect
recovery above is a property of a generator we constructed and does not
transfer to causal discovery on observational data: the probes do not measure
a causal mechanism, and success under a controlled generator does not
establish causal discovery on real data.  We therefore describe
Phase~III as a mechanism \emph{diagnostic} throughout, and it should guide
investigation without justifying a corrective action on its own.

\paragraph{Two-disease validation cohort}
The validation run reproduces both patterns at smaller scale: mixed mechanisms
split 4 model-driven and 2 label-noise-driven across 6 runs with no
no-material-disparity verdicts, and the misspecified proxy returned
no-material-disparity in 5 of 6 with mean CF excess $-0.249$.  The direction
and the silence of the misspecification failure are consistent across cohorts.

\subsection{Phase V drift monitoring across cohorts}
\label{sec:monitoring}

\begin{figure}[t]
\centering
\includegraphics[width=0.9\columnwidth]{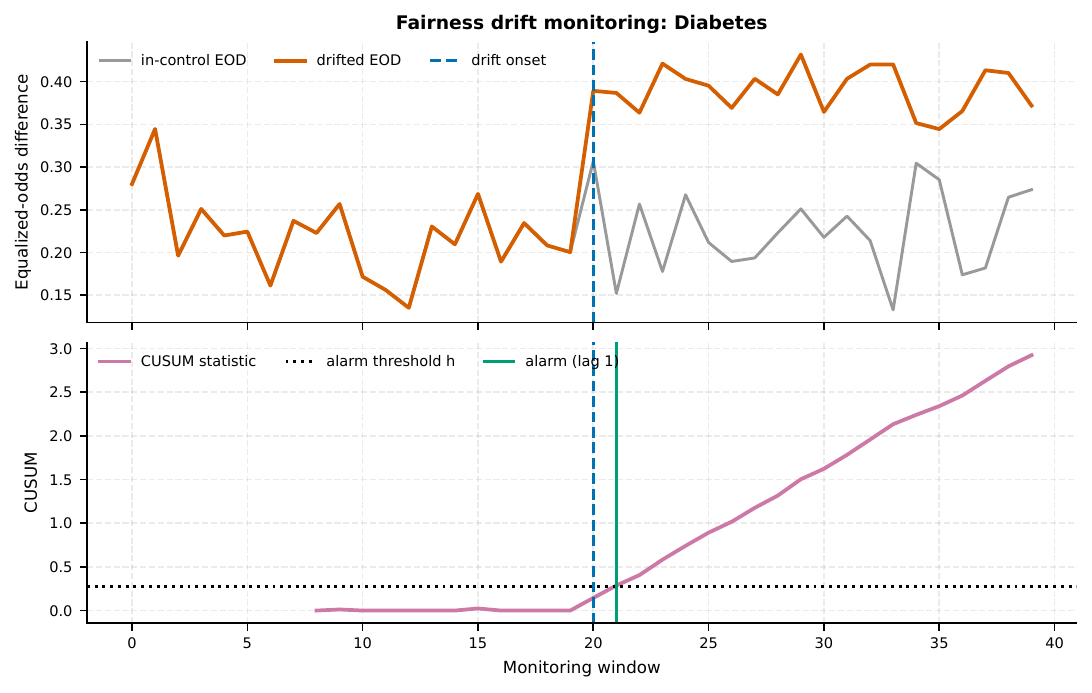}
\caption{Representative EOD stream and one-sided CUSUM trace.  The vertical
  line marks injected drift onset; the alarm is the first threshold crossing
  after onset.}
\label{fig:cusum}
\end{figure}

The CUSUM monitor tracks the EOD stream for the composite Medicaid/uninsured
population, roughly 30\% of the cohort, against commercially insured patients.
Table~\ref{tab:cusum_sweep} sweeps the alarm threshold over $4\sigma$,
$6\sigma$, and $8\sigma$ across all 48 disease-seed runs, reporting detection,
delay, and in-control average run length; Fig.~\ref{fig:cusum} shows a
representative stream and its CUSUM trace.

\begin{table}[t]
\caption{CUSUM threshold sweep across 16 diseases $\times$ 3 seeds (48 runs
  per threshold; $\kappa{=}0.005$, $W{=}1{,}200$, shift injected at
  $t^*{=}20$, 40 windows per run, 12 monitored in-control windows per run).
  The label is the multiple of the in-control SD used to set $h$ adaptively.
  Lag is conditional on detection, so the columns are not comparable without
  the detection count beside them; the median is given because the
  distribution is right-skewed.  $\text{ARL}_0$ is the in-control average run
  length in windows.  ``FA runs'' counts how many of the 48 runs produced at
  least one false alarm.  Source:
  \texttt{outputs/sdoh\_multidisease/monitoring\_sweep.csv}.}
\label{tab:cusum_sweep}
\centering\footnotesize
\setlength{\tabcolsep}{3pt}
\resizebox{\columnwidth}{!}{%
\begin{tabular}{lcccccc}
\toprule
\textbf{Thresh.} & \textbf{Detected} & \textbf{Lag (mean)} &
  \textbf{Lag (med.)} & \textbf{False alarms} & \textbf{FA runs} &
  $\boldsymbol{\text{ARL}_0}$ \\
\midrule
$4\sigma$ & 41/48 & 1.46 & 0 & 59 & 11/48 & 10 \\
$6\sigma$ & 40/48 & 2.50 & 1 & 27 & 5/48  & 21 \\
$8\sigma$ & 38/48 & 2.87 & 2 & 11 & 3/48  & 52 \\
\bottomrule
\end{tabular}%
}
\end{table}

At the $6\sigma$ reference setting the monitor detects 40 of 48 injected
shifts with a mean lag of 2.50 windows, roughly 3{,}000 patients at
$W=1{,}200$, and produces 27 false alarms.  Raising the threshold from
$4\sigma$ to $8\sigma$ reduces false alarms more than five-fold and lifts
$\text{ARL}_0$ from 10 to 52 windows, and this is the only quantity that
improves: detections fall from 41 to 38 and mean lag among detected runs rises
from 1.46 to 2.87 windows, an increase from roughly 1{,}800 to 3{,}400
patients scored by a degraded model before any alarm.  The lag figure
understates the cost of a strict threshold because it conditions on detection,
so the runs a stricter threshold loses entirely drop out of the average instead
of entering it as an unbounded delay.  No threshold detected every
run.

Where the errors fall is the more useful finding.  Both failure modes separate
almost perfectly by cohort realization, and barely at all by disease.  At
$6\sigma$,
all 27 false alarms come from seed 37, concentrated in 5 of its 16 runs, while
seeds 11 and 23 produce none at all; conversely 7 of the 8 missed detections
are seed 23, whose per-seed detection rate is 9 of 16 against 15 of 16 for
seed 11 and 16 of 16 for seed 37 ($\chi^2$ $p=0.002$).  Seven runs go
undetected at every threshold in the sweep, and all seven are seed 23.  The
mechanism is visible in the adaptive threshold itself: seed 23 has the
noisiest in-control stream ($\hat{\sigma}_0=0.054$ on average against $0.048$
for seed 11), so the adaptive rule hands it the highest alarm threshold
($h=0.33$ against $0.29$) precisely in the cohort where the shift is hardest
to see, while seed 37's lower and more variable in-control mean
($\mu_0=0.190\pm0.041$ against $0.222\pm0.017$) makes it prone to spurious
crossings.

This changes the practical reading of the sweep.  Within a cohort the two
error types barely coexist: the cohort that misses almost never false-alarms,
and the cohort that false-alarms never misses.  What looks like a
threshold-tuning trade-off is largely a transfer problem.  An operator who
calibrates $h$ on one in-control period and deploys it on a population whose
baseline EOD variance differs will land in a different error regime from the
one they tuned for, and the adaptive rule meant to protect against exactly
this works against detection when the baseline is noisy.  We therefore report the sweep and recommend no
default, noting that a principled choice requires what this benchmark does
not supply: a target in-control false-alarm rate, equivalently a target
$\text{ARL}_0$, fixed in advance and calibrated on in-control data from the
deployment population itself, with re-calibration when that population
changes.

\paragraph{Two-disease validation cohort}
The same sweep on the validation run (diabetes and sepsis, 3 seeds each; 6
runs per threshold) shows the same monotone structure: detections held at 4 of
6 at every threshold, mean lag among detected runs rose from 1.25 to 1.75 to
5.25 windows, and false alarms fell from 6 to 4 to 0.  The two missed runs
were missed at all three thresholds, which points to a stream whose noise
level defeats the monitor at any setting.  The
false-alarm reduction is steeper and the lag penalty larger in this cohort,
which is why the 48-run sweep supplies the quantitative claims above.  The
$8\sigma$ setting reaching zero false alarms in this cohort is worth stating
plainly as a caution: zero false alarms is not automatically the best choice,
because the trade-off for a high threshold is a monitor delayed long enough
that a degraded model keeps scoring patients in the interval.  The operating
threshold has to be chosen from the clinical cost of that delay weighed
against the review burden a spurious alarm imposes.  That is a
deployment-specific judgment, and no statistic settles it.

\subsection{Audit priorities by decision context}
\label{sec:usecases_results}

Table~\ref{tab:usecases} summarizes the three decision-support contexts of
\S\ref{sec:clinical_context}.  The priority metric follows from the asymmetry
between the two error costs, not from which metric is most convenient to
compute, and the three tasks do not resolve to the same choice.

For \emph{30-day readmission} and \emph{sepsis early warning} the dominant
harm is a missed case, so the priority metric is the equalized missed-case
rate.  Because the false-negative rate is the complement of the true-positive
rate, equalizing one equalizes the other; we report the false-negative form so
that the equalized quantity is the harm itself.  The two tasks differ in
tolerance for false alarms.  Readmission alerts consume case-management
capacity over days, so a moderate false-alarm burden is absorbable, whereas
sepsis alerts fire into a workflow where excess alarms degrade response to
genuine ones, so the missed-case constraint must be met without inflating the
alarm rate.  The stakes in the sepsis case are set by the time dependence of
the underlying harm~\cite{kumar2006,seymour2017,wong2021}.

For \emph{in-hospital mortality prediction} the score is consumed as a
probability and not as a binary flag, so rank-order agreement will not do.  A model that systematically overestimates mortality for one
subgroup can contribute to disproportionate care limitation even when its
per-group AUC is acceptable~\cite{sjoding2020,barnato2007}.
Group-stratified ECE is the priority metric, supplemented by MCE, and
\S\ref{sec:calibration} shows this supplement is not cosmetic: mean MCE is
about five times mean ECE and ranks the axes differently ($\rho=0.65$), so the
two can disagree about which subgroup is worst served.

The ordering says which metric to privilege when metrics conflict, and it has
a direct consequence for the audit.  In this benchmark the most disparate
metric is often not the one a given task depends on, so a single-metric audit
can return a clean result on the axis that matters least.

\begin{table}[t]
\caption{Clinical decision-support context for three representative tasks.
  Metric priority follows from the asymmetric cost of the dominant error type.
  FNR and TPR are complementary; the false-negative form is reported so that
  the equalized quantity is the harm itself.}
\label{tab:usecases}
\centering\footnotesize
\begin{tabular}{p{1.45cm}p{1.5cm}p{1.6cm}p{1.75cm}}
\toprule
\textbf{Task} & \textbf{Users} & \textbf{Timing} & \textbf{Priority metric} \\
\midrule
30-day readmission &
  Discharge planners; case managers; post-acute coordinators &
  Final days of index admission &
  Equalized missed-case rate (FNR) \\[4pt]
Sepsis early warning &
  Bedside nurses; rapid-response teams; ED/ICU physicians &
  Continuous during ED/ICU stay &
  Equalized missed-case rate (FNR), at fixed alarm rate \\[4pt]
In-hospital mortality &
  ICU attendings; critical-care teams &
  Continuous during ICU stay (hourly in some models) &
  Group-stratified ECE, with MCE \\
\bottomrule
\end{tabular}
\end{table}

% ============================================================
%  VIII. DISCUSSION
% ============================================================
\section{Discussion}

\subsection{What generalizes}

The numeric results in this paper are properties of a generator we wrote, and
none of them estimates disparity in a clinical population.  Three structural
claims about the audit itself should survive the move to real data.

The first concerns silence.  A no-material-disparity verdict from Phase~III
comes out the same whether the model is clean or the probe was pointed at the
wrong columns (\S\ref{sec:attribution}); a non-significant axis in Phase~II
comes out the same whether the disparity is absent or the smallest cell simply
cannot resolve it (\S\ref{sec:evidence}).  One cheap remedy covers both cases:
print the instrument's resolution next to its reading, which means the proxy
set for Phase~III and $n^{+}_{\min}$, MDE, and $R$ for Phase~II.

The second is that metric improvements fail to compose.  Calibration and
error-rate parity pull in different directions, and \S\ref{sec:mitigation}
gives a run-level instance of the impossibility
results~\cite{chouldechova2017,kleinberg2016}, with the better calibrator
turning out to be the unreliable fairness intervention.  A single aggregate
fairness score would have buried this, which is why we report a metric vector.

The third concerns aggregation.  Averaging mitigation results across runs made
Platt scaling look like a mild EOD regression, when the defensible finding is
an effect centered near zero with a spread several times larger.  Applied to
Phase~V, the same averaging would have reported a detection/false-alarm
trade-off that no single cohort ever experiences.

\subsection{Limitations}
\label{sec:limitations}

The framework is evaluated on a synthetic benchmark with controlled injected
disparities.  We chose that deliberately, since controlled generation supports recovery
tests and lets us exercise each phase independently.  The generator also
encodes the relationships among SDOH attributes, outcomes, and proxy features,
which makes the observed ranking of axes a benchmark property and not an
estimate of clinical disparity.  In real EHR data the data-generating
process is unknown, the sensitive attribute is often missing or poorly coded,
and label noise is inseparable from documentation practice.

Statistical scope is limited in three specific ways.  Inference is
prespecified for the 15 single axes on the HGBT model only, which leaves the
three intersections carrying the largest point estimates in the benchmark and
no evidence at all (\S\ref{sec:evidence}).  No subgroup cell reaches the
support the power analysis calls for (\S\ref{sec:support}), and 12 of the 15
tested axes have $R<1$.  The significance pattern we report is also
conditional on three seed cohorts, and \S\ref{sec:monitoring} shows how far a
single cohort realization can move a result.

The Phase~IV mitigation is post-hoc.  Inverse-propensity weighting balances
measured feature distributions on the calibration set but does not identify
causal effects or address unmeasured confounding.  The threshold search is
coordinate descent and returns a local optimum, so the reported EOD reductions
are lower bounds under the accuracy constraint.  Group-specific thresholds
require group membership at inference time and therefore explicit legal,
ethical, and clinical governance, and selection on a calibration split does
not guarantee improvement on future data.

A broader limitation is that optimizing one fairness metric gives no guarantee
about the others.  Threshold optimization was highly
effective on EOD, reducing it in all 48 runs, but it acts only on the decision
boundary and carries no calibration objective of its own: the ECE gain
reported beside it is inherited from the Platt stage that precedes it, not
produced by the thresholds.  Platt scaling improved probability calibration
but was not a successful fairness intervention.  Post-hoc calibration can
therefore increase disparity in error rates, so the mitigation has to be
matched to the quantity the deployment actually depends on
(\S\ref{sec:clinical_context}), which is often the harder one to improve.

Intersectional fairness is evaluated for three axis pairs and not
exhaustively.  Intersectional strata consistently show amplified EOD relative
to single-axis strata, reaching $0.703\pm0.031$ on race~$\times$~insurance
against $0.509\pm0.060$ for insurance alone.  Compound disadvantage of this
kind is well documented in the SDOH literature.  Full combinatorial
intersectional analysis, with a prespecified minimum cell count and an
inference protocol extended to those strata, is the direct next step.

\subsection{Outlook}

Real-world validation remains necessary before any deployment-level
conclusion.  The recommended next steps are: (i)~replication on MIMIC-IV
readmission and mortality tasks~\cite{johnson2023} with race and insurance as
sensitive attributes, reporting $n^{+}_{\min}$ and $R$ per axis;
(ii)~longitudinal evaluation on time-stamped admission data to test Phase~V
against observed drift instead of injected drift, with $h$ calibrated to a
prespecified $\text{ARL}_0$ on a held-out in-control period; (iii)~a
Phase~III protocol in which the proxy set is registered in advance and the
diagnostic reports a sensitivity range over plausible proxy sets in place of a
single label; and (iv)~a prospective audit protocol designed with a clinical
partner.

% ============================================================
%  IX. CONCLUSION
% ============================================================
\section{Conclusion}

KAISEN separates subgroup measurement, confirmatory testing, mechanism
diagnostics, post-hoc mitigation, and longitudinal monitoring into a
reproducible workflow, and we pushed each component until it broke.  In the
controlled 16-disease benchmark, aggregate performance conceals large subgroup
error-rate gaps, and which of those gaps reaches significance depends on
effect size measured against each axis's own detectable floor, with $\rho$
rising from $0.56$ to $0.78$ under standardization.  Threshold optimization
reduces EOD in 48 of 48 held-out runs (paired $\Delta=-0.285$, CI
$[-0.313,-0.252]$), while group-wise Platt scaling calibrates better and
performs like a coin flip on EOD.  The mechanism diagnostic classifies all 144
controlled cases and none of 48 misspecified-proxy cases, giving no signal
that it has failed.  The drift monitor's detection failures and false alarms
follow the cohort realization, so a threshold tuned on one in-control period
fails to transfer.

Every one of these components fails quietly, and the same fix applies to all
of them: publish the instrument's resolution beside its reading.  In practice
that means the subgroup support behind a $p$-value and the proxy set behind a
no-material-disparity verdict, together with the per-run spread behind any
mitigation mean and the in-control ARL behind any detection rate.  Fairness
metrics belong in a vector with uncertainty across all relevant subgroups,
because recalibration and equalized-odds objectives need separate
interventions and progress on one buys nothing on the other.  Mitigation and
diagnostic claims should be validated on untouched held-out data.  Clinical
and causal conclusions need external evidence well beyond this synthetic
benchmark.

% ============================================================
%  REPRODUCIBILITY
% ============================================================
\section*{Reproducibility}
\label{sec:reproducibility}

A script emits every quantitative claim in this manuscript from a committed
artifact; none was transcribed by hand.
\texttt{scripts/generate\_paper\_results.py} writes the point estimates, and
\texttt{scripts/generate\_extended\_results.py} the derived statistics of
\S\ref{sec:support}, \S\ref{sec:evidence}, \S\ref{sec:calibration},
\S\ref{sec:mitigation}, and \S\ref{sec:monitoring}, both as LaTeX macro files
alongside this source.  Experiment runs record a git commit, a SHA-256 hash of
the configuration, the Python version, and the platform in
\texttt{run\_metadata.json}; the results reported here correspond to commit
\texttt{46341ee0} and config hash \texttt{4bb87ed0}.  Seeds are fixed at
$\{11,23,37\}$ and all models are deterministic given a seed.  The benchmark
generator, the audit implementation, the 16-disease and 2-disease output
artifacts, and the figure-generation code are released
at~\cite{kaiser_repo}.\footnote{The public repository and Python package
retain the project's original \texttt{KAISER} identifier.}

% ============================================================
%  AUTHOR CONTRIBUTIONS
% ============================================================
\section*{Author Contributions}

S.~Roy designed the audit framework and synthetic benchmark, implemented
Phases I--V, performed the support and evidence analysis of
\S\ref{sec:support}--\S\ref{sec:evidence} and the calibration robustness
analysis of \S\ref{sec:calibration}, and wrote the manuscript.  N.~Chavan
characterized the three clinical decision-support contexts of
\S\ref{sec:clinical_context} and \S\ref{sec:usecases_results} and the
deployed-system disparity examples of Table~\ref{tab:disparity_examples}.
S.~Girmachew analyzed the mitigation reliability (\S\ref{sec:mitigation}),
CUSUM threshold sensitivity (\S\ref{sec:monitoring}), and attribution stress
tests (\S\ref{sec:attribution}), including the two-disease validation-cohort
replications reported in each.  All authors reviewed and approved the final
manuscript. Large Language Models were used to improve prose and clarity in writing. All information, results, and models were the authors own.

% ============================================================
%  AUTHORSHIP STATEMENT
% ============================================================
\section*{Use of AI Tools and Authorship Statement}

The authors used generative AI tools during software development and
manuscript preparation.  Assistance included code review, test generation,
debugging, documentation, prose revision, and consistency checks between
experiment artifacts and the manuscript.  The authors specified the research
questions, reviewed the implementation, generated the artifacts, decided which
analyses and claims to retain, and take responsibility for the accuracy,
originality, and integrity of the work.

% ============================================================
%  REFERENCES
% ============================================================


\begin{thebibliography}{42}

\bibitem{dwork2012}
C.~Dwork, M.~Hardt, T.~Pitassi, O.~Reingold, and R.~Zemel,
``Fairness through awareness,''
in \textit{Proc.\ 3rd Innov.\ Theor.\ Comput.\ Sci.\ Conf.}, 2012, pp.~214--226.

\bibitem{hardt2016}
M.~Hardt, E.~Price, and N.~Srebro,
``Equality of opportunity in supervised learning,''
in \textit{Adv.\ Neural Inf.\ Process.\ Syst.}, vol.~29, 2016.

\bibitem{chouldechova2017}
A.~Chouldechova,
``Fair prediction with disparate impact: A study of bias in recidivism
prediction instruments,''
\textit{Big Data}, vol.~5, no.~2, pp.~153--163, 2017.

\bibitem{barocas2019}
S.~Barocas, M.~Hardt, and A.~Narayanan,
\textit{Fairness and Machine Learning: Limitations and Opportunities}.
fairmlbook.org, 2019.

\bibitem{kleinberg2016}
J.~Kleinberg, S.~Mullainathan, and M.~Raghavan,
``Inherent trade-offs in the fair determination of risk scores,''
2016, arXiv:1609.05807.

\bibitem{obermeyer2019}
Z.~Obermeyer, B.~Powers, C.~Vogeli, and S.~Mullainathan,
``Dissecting racial bias in an algorithm used to manage the health of
populations,''
\textit{Science}, vol.~366, no.~6464, pp.~447--453, 2019.

\bibitem{sjoding2020}
M.~W.~Sjoding, R.~P.~Dickson, T.~J.~Iwashyna, S.~E.~Gay, and T.~S.~Valley,
``Racial bias in pulse oximetry measurement,''
\textit{New England J.\ Med.}, vol.~383, no.~25, pp.~2477--2478, 2020.

\bibitem{oakden2020}
L.~Oakden-Rayner, J.~Dunnmon, G.~Carneiro, and C.~R\'{e},
``Hidden stratification causes clinically meaningful failures in machine
learning for medical imaging,''
in \textit{Proc.\ ACM Conf.\ Health, Inference, Learning}, 2020, pp.~151--159.

\bibitem{hosmer1980}
D.~W.~Hosmer and S.~Lemeshow,
``Goodness of fit tests for the multiple logistic regression model,''
\textit{Commun.\ Statist.--Theory Methods}, vol.~9, no.~10, pp.~1043--1069, 1980.

\bibitem{nixon2019}
J.~Nixon, M.~W.~Dusenberry, L.~Zhang, G.~Jerfel, and D.~Tran,
``Measuring calibration in deep learning,''
in \textit{CVPR Workshops}, 2019.

\bibitem{delong1988}
E.~R.~DeLong, D.~M.~DeLong, and D.~L.~Clarke-Pearson,
``Comparing the areas under two or more correlated receiver operating
characteristic curves: a nonparametric approach,''
\textit{Biometrics}, pp.~837--845, 1988.

\bibitem{kusner2017}
M.~J.~Kusner, J.~Loftus, C.~Russell, and R.~Silva,
``Counterfactual fairness,''
in \textit{Adv.\ Neural Inf.\ Process.\ Syst.}, vol.~30, 2017.

\bibitem{kamiran2012}
F.~Kamiran and T.~Calders,
``Data preprocessing techniques for classification without discrimination,''
\textit{Knowledge Inf.\ Syst.}, vol.~33, no.~1, pp.~1--33, 2012.

\bibitem{zhang2018}
B.~H.~Zhang, B.~Lemoine, and M.~Mitchell,
``Mitigating unwanted biases with adversarial learning,''
in \textit{Proc.\ AAAI/ACM Conf.\ AI, Ethics, Society}, 2018, pp.~335--340.

\bibitem{guo2017}
C.~Guo, G.~Pleiss, Y.~Sun, and K.~Q.~Weinberger,
``On calibration of modern neural networks,''
in \textit{Int.\ Conf.\ Mach.\ Learn.}, 2017, pp.~1321--1330.

\bibitem{page1954}
E.~S.~Page,
``Continuous inspection schemes,''
\textit{Biometrika}, vol.~41, no.~1/2, pp.~100--115, 1954.

\bibitem{montgomery2009}
D.~C.~Montgomery,
\textit{Introduction to Statistical Quality Control}, 6th~ed.
Wiley, 2009.

\bibitem{giobergia2025}
F.~Giobergia, E.~Pastor, L.~de Alfaro, and E.~Baralis,
``Detecting interpretable subgroup drifts,''
in \textit{Proc.\ 31st ACM SIGKDD Conf.\ Knowledge Discovery and Data Mining},
2025, doi:~10.1145/3690624.3709259.

\bibitem{johnson2023}
A.~Johnson, L.~Bulgarelli, L.~Shen, A.~Gayles, A.~Shammout, S.~Horng,
T.~J.~Pollard, L.~A.~Celi, and R.~Mark,
``MIMIC-IV, a freely accessible electronic health record dataset,''
\textit{Scientific Data}, vol.~10, no.~1, p.~1, 2023.

\bibitem{bh1995}
Y.~Benjamini and Y.~Hochberg,
``Controlling the false discovery rate: A practical and powerful approach to
multiple testing,''
\textit{J.\ Roy.\ Statist.\ Soc.\ Ser.\ B}, vol.~57, no.~1, pp.~289--300, 1995.

\bibitem{efron1994}
B.~Efron and R.~J.~Tibshirani,
\textit{An Introduction to the Bootstrap}.
Chapman \& Hall, 1994.

\bibitem{good2005}
P.~I.~Good,
\textit{Permutation, Parametric and Bootstrap Tests of Hypotheses}, 3rd~ed.
Springer, 2005.

\bibitem{sklearn2011}
F.~Pedregosa et al.,
``Scikit-learn: Machine learning in Python,''
\textit{J.\ Mach.\ Learn.\ Res.}, vol.~12, pp.~2825--2830, 2011.

\bibitem{hp2030}
U.S.\ Department of Health and Human Services,
``Healthy People 2030: Social Determinants of Health,''
Office of Disease Prevention and Health Promotion, 2020.
[Online]. Available:
\url{https://health.gov/healthypeople/priority-areas/social-determinants-health}

\bibitem{kind2014}
A.~J.~Kind, S.~Jencks, J.~Brock, M.~Yu, W.~Bartels, L.~Ehlenbach,
C.~Greenberg, and M.~Smith,
``Neighborhood socioeconomic disadvantage and 30-day rehospitalization,''
\textit{Ann.\ Internal Med.}, vol.~161, no.~11, pp.~765--774, 2014.

\bibitem{decker2011}
S.~L.~Decker,
``Medicaid physician fees and the quality of medical care of Medicaid patients
in the USA,''
\textit{Review of Economics of the Household}, vol.~9, no.~4, pp.~497--514, 2011.

\bibitem{seymour2016}
C.~W.~Seymour, V.~X.~Liu, T.~J.~Iwashyna, F.~M.~Brunkhorst, T.~D.~Rea,
A.~Scherag, G.~Rubenfeld, G.~J.~Kahn, C.~S.~Shankar-Hari, M.~Singer,
C.~S.~Deutschman, G.~S.~Escobar, and D.~C.~Angus,
``Assessment of clinical criteria for sepsis: for the third international
consensus definitions for sepsis and septic shock (Sepsis-3),''
\textit{JAMA}, vol.~315, no.~8, pp.~762--774, 2016.

\bibitem{seyyed_kalantari2021}
L.~Seyyed-Kalantari, H.~Zhang, M.~B.~A.~McDermott, I.~Y.~Chen, and
M.~Ghassemi,
``Underdiagnosis bias of artificial intelligence algorithms applied to chest
radiographs in under-served patient populations,''
\textit{Nature Medicine}, vol.~27, no.~12, pp.~2176--2182, 2021.

\bibitem{kumar2006}
A.~Kumar, D.~Roberts, K.~E.~Wood, B.~Light, J.~E.~Parrillo, S.~Sharma,
R.~Suppes, D.~Feinstein, S.~Zanotti, L.~Taiberg, D.~Gurka, A.~Kumar, and
M.~Cheang,
``Duration of hypotension before initiation of effective antimicrobial
therapy is the critical determinant of survival in human septic shock,''
\textit{Critical Care Medicine}, vol.~34, no.~6, pp.~1589--1596, 2006.

\bibitem{seymour2017}
C.~W.~Seymour, F.~Gesten, H.~C.~Prescott, M.~E.~Friedrich, T.~J.~Iwashyna,
G.~S.~Phillips, S.~Lemeshow, T.~Osborn, K.~M.~Terry, and M.~M.~Levy,
``Time to treatment and mortality during mandated emergency care for
sepsis,''
\textit{New England Journal of Medicine}, vol.~376, no.~23,
pp.~2235--2244, 2017.

\bibitem{wong2021}
A.~Wong, E.~Otles, J.~P.~Donnelly, A.~Krumm, J.~McCullough,
O.~DeTroyer-Cooley, J.~Pestrue, M.~Phillips, J.~Konye, C.~Penoza,
M.~Ghous, and K.~Singh,
``External validation of a widely implemented proprietary sepsis
prediction model in hospitalized patients,''
\textit{JAMA Internal Medicine}, vol.~181, no.~8, pp.~1065--1070, 2021.

\bibitem{haas2022}
R.~Haas and S.~C.~McGill,
\textit{Artificial Intelligence for the Prediction of Sepsis in Adults},
CADTH Horizon Scan.
Ottawa, ON, Canada: Canadian Agency for Drugs and Technologies in Health,
Mar.\ 2022.

\bibitem{papareddy2025}
P.~Papareddy, T.~J.~Lobo, M.~Holub, H.~Bouma, J.~Maca, N.~Strodthoff, and
H.~Herwald,
``Transforming sepsis management: AI-driven innovations in early detection and
tailored therapies,''
\textit{Critical Care}, vol.~29, art.~366, 2025.

\bibitem{mazza2026}
G.~Mazza et al.,
``Artificial intelligence models for mortality and outcome prediction in
intensive care unit sepsis: A systematic review,''
\textit{Journal of Personalized Medicine}, vol.~16, no.~7, art.~346, 2026.

\bibitem{kansagara2011}
D.~Kansagara, H.~Englander, A.~Salanitro, D.~Kagen, C.~Theobald,
M.~Freeman, and S.~Kripalani,
``Risk prediction models for hospital readmission: a systematic review,''
\textit{JAMA}, vol.~306, no.~15, pp.~1688--1698, 2011.

\bibitem{escobar2020}
G.~J.~Escobar, V.~X.~Liu, A.~Schuler, B.~Lawson, J.~D.~Greene, and
P.~Kipnis,
``Automated identification of adults at risk for in-hospital clinical
deterioration,''
\textit{New England Journal of Medicine}, vol.~383, no.~20,
pp.~1951--1960, 2020.

\bibitem{knaus1985}
W.~A.~Knaus, E.~A.~Draper, D.~P.~Wagner, and J.~E.~Zimmerman,
``APACHE II: a severity of disease classification system,''
\textit{Critical Care Medicine}, vol.~13, no.~10, pp.~818--829, 1985.

\bibitem{vincent1996}
J.-L.~Vincent, R.~Moreno, J.~Takala, S.~Willatts, A.~De Mendonca,
H.~Bruining, C.~K.~Reinhart, P.~M.~Suter, and L.~G.~Thijs,
``The SOFA (sepsis-related organ failure assessment) score to describe
organ dysfunction/failure,''
\textit{Intensive Care Medicine}, vol.~22, no.~7, pp.~707--710, 1996.

\bibitem{raith2017}
E.~P.~Raith, A.~A.~Udy, M.~Bailey, S.~McGloughlin, C.~MacIsaac,
R.~Bellomo, and D.~V.~Pilcher,
``Prognostic accuracy of the SOFA score, SIRS criteria, and qSOFA score
for in-hospital mortality among adults with suspected infection admitted
to the intensive care unit,''
\textit{JAMA}, vol.~317, no.~3, pp.~290--300, 2017.

\bibitem{kipnis2016}
P.~Kipnis, B.~J.~Turk, D.~A.~Wulf, J.~C.~LaGuardia, V.~X.~Liu,
M.~M.~Churpek, S.~Romero-Brufau, and G.~J.~Escobar,
``Development and validation of an electronic medical record-based alert
score for detection of inpatient deterioration outside the ICU,''
\textit{Journal of Biomedical Informatics}, vol.~64, pp.~10--19, 2016.

\bibitem{barnato2007}
A.~E.~Barnato, C.-C.~H.~Chang, O.~Saynina, and A.~M.~Garber,
``Influence of race on inpatient treatment intensity at the end of life,''
\textit{Journal of General Internal Medicine}, vol.~22, no.~3, pp.~338--345,
2007.

\bibitem{kaiser_repo}
S.~Roy, S.~Girmachew, and N.~Chavan,
``KAISEN: Clinical AI Fairness Audit Framework,'' 2024.
[Online]. Available: \url{https://github.com/ARC-Labs-group/KAISER}

\end{thebibliography}
\end{document}